\documentclass[times, 10pt]{elsarticle}
\usepackage{graphicx} 
\usepackage{amssymb}
\usepackage{amsmath}
\usepackage[english]{babel}
\usepackage[colorlinks]{hyperref}
\usepackage{booktabs}
\usepackage[hang,raggedright,tight]{subfigure}
\usepackage{comment}
\usepackage{multirow}
\usepackage{units}
\journal{Pattern Recognition}

\usepackage{colortbl} 
\usepackage{xcolor}

\newcommand{\TP}{\mathrm{TP}}
\newcommand{\TN}{\mathrm{TN}}
\newcommand{\FP}{\mathrm{FP}}
\newcommand{\FN}{\mathrm{FN}}

\newcommand{\FPR}{\mathrm{FPR}}
\newcommand{\FNR}{\mathrm{FNR}}

\usepackage{enumitem}

\newcommand{\labelA}{\rotatebox{90}{\parbox{1.7cm}{\centering age only}}}
\newcommand{\labelB}{\rotatebox{90}{\parbox{1.7cm}{\centering age + under-18}}}
\newcommand{\labelC}{\rotatebox{90}{\parbox{1.7cm}{\centering age + 12,15,18,21}}}

\usepackage[absolute]{textpos}
\begin{document}

\begin{textblock*}{20cm}(1cm,1cm)
© 2025. This manuscript version is made available under the  \href{https://creativecommons.org/licenses/by-nc-nd/4.0/}{CC-BY-NC-ND 4.0 license}.
\end{textblock*}

\date{November 2025}

\begin{frontmatter}

\title{Underage Detection through a Multi-Task and MultiAge Approach for Screening Minors in Unconstrained Imagery}

\author{Christopher Gaul\corref{cor1}}
\ead{cgau@unileon.es}
\cortext[cor1]{Corresponding author}

\author{Eduardo Fidalgo}
\author{Enrique Alegre}
\author{Rocío Alaiz Rodríguez}
\author{Eri Pérez Corral}

\address{Department of Electrical, Systems and Automation Engineering, Universidad de León, León, Spain}

\begin{abstract}
Accurate automatic screening of minors in unconstrained images requires models robust to distribution shift and resilient to the under-representation of children in public datasets. 
To address these issues, we propose a multi-task architecture with dedicated under/over-age discrimination tasks based on a frozen FaRL vision-language backbone joined with a compact two-layer MLP that shares features across one age-regression head and four binary underage heads (12, 15, 18, and 21 years). This design focuses on the legally critical age range while keeping the backbone frozen.
Class imbalance is mitigated through an $\alpha$-reweighted focal loss and age-balanced mini-batch sampling,
while an age gap removes ambiguous samples near thresholds.

Evaluation is conducted on our new Overall Underage Benchmark (303k cleaned training images, 110k test images), defining both the ``ASORES-39k'' restricted overall test, which removes the noisiest domains, and the age estimation wild-shifts test ``ASWIFT-20k'' of 20k-images, stressing extreme poses ($>$45°), expressions, and low image quality to emulate real-world shifts.

Trained on the cleaned overall set with resampling and age gap, our multiage model ``F'' reduces the mean absolute error on  ASORES-39k from \unit[4.175]{y} (age-only baseline) to \unit[4.068]{y} and improves under-18 detection from F2 score of $0.801$ to $0.857$ at $1\%$ false-adult rate.
Under the ASWIFT-20k, the same configuration nearly sustains $0.99$ recall
while F2 rises from $0.742$ to $0.833$, demonstrating robustness to domain shift.
\end{abstract}


\end{frontmatter}

\section{Introduction}

Facial age estimation aims to predict a person's age from a facial image \cite{Angulu_2018_survey}.
Although it is a regression task, it is often formulated as classification, by rounding ages or defining age groups \cite{Eidinger_2014_Adience}.
The binary case, distinguishing subjects below or above a threshold, is known as age verification or,
conversely, as underage or minor detection.

Age estimation and underage detection are increasingly important across a wide range of applications,
including age-restricted services, targeted advertising, social media moderation, and, most critically, the protection of minors.
Applications involve equipping devices such as gambling or cigarette vending machines, or age-restricted smartphone applications with automated facial age-verification systems. 
Minor detection is essential in enforcing age-related regulations in digital environments and to combat the distribution of Child Sexual Exploitation Material (CSEM).
CSEM hidden in huge amounts of data is detected in a two-step process, consisting of the detection of sexual activity \cite{Gangwar_2024_DeepHSAR} and the detection of underage subjects.

For CSEM detection, the costs of classification errors are asymmetric: a false adult detection means an undetected case of child abuse, while a false minor detection only means that the officer who inspects the detections will find out it is a false detection.
Therefore, such systems are typically operated at high recall values, (e.g., 0.99)— to avoid missing minors — and at lower precision (accepting some false alarms).

While significant progress has been made in facial age estimation, most existing solutions are designed for controlled settings and tend to perform well when image conditions are ideal.
However, real-world scenarios (``in the wild") present substantial challenges, including variations in head pose, facial expressions, lighting conditions, image resolution, and frequent occlusions \cite{Jeuland_2022_occlusion}.
These factors severely impact the reliability of the model and often degrade its performance.
These challenges become even more pronounced when distinguishing between children, adolescents, and adults. 
This is precisely where accurate detection becomes more critical for legal and safety considerations.
Furthermore, publicly available facial datasets, like MORPH \cite{Ricanek_2006_MORPH}, AFAD \cite{Niu_2016_AFAD} or AgeDB \cite{Moschoglou_2017_AgeDB}, have skewed distributions with a significant under-representation of children and adolescents. This imbalance leads to lower performance for younger age groups. Consequently, the effectiveness of underage detection systems is compromised in critical applications, such as CSEM investigations.
%
%
In this work, we address the challenges of robust facial age estimation and underage detection with three objectives:
\begin{itemize}
  \item A \emph{benchmark} for age-estimation models under unconstrained conditions.
  \item Improving age-estimation models, with focus on such \emph{unconstrained} scenarios.
  \item Enhancing age estimation \emph{for minors} and the discrimination of minors and adults.
\end{itemize}
To achieve these, we propose a comprehensive framework for unconstrained environments. The main contributions of this study are as follows:
\begin{itemize}
  \item The Overall Underage benchmark and its restricted subset ASORES-39k, designed for the accurate evaluation of age estimation and underage detection models while excluding examples with noisy or unreliable labels.
  \item ASWIFT-20k, a novel \emph{wild test} benchmark, for robustness evaluation in diverse and challenging imaging conditions, including variations in pose, facial expression, illumination, and image quality. 
  \item A multitask learning architecture that combines age regression with dedicated binary classifiers for under/over-age thresholds, enhancing discriminative focus where it is most critical.
  \item An age-balanced sampling strategy during training to mitigate class imbalance. This ensures equal representation of minors, enhancing the ability of the model to generalize across underrepresented age groups.
  \item We also explore $\alpha$-reweighted focal loss to mitigate the effects of class imbalance and to prioritize difficult and minority-class samples
\end{itemize}
The proposed methodology aims to improve facial age estimation, particularly to distinguish between prepubescent, pubescent, and adult individuals. Based on our experience working with law enforcement agencies (LEAs) on previous projects and given the same age for males and females, we established the relevant ages as 12 (prepubescent), 15 (pubescent), and 18 or 21 (adult) \cite{Blanchard2009}.
The focus is on developing robust underage detection systems that perform well across varied real-world conditions, including diverse poses, facial expressions, and poor image quality. In addition, it addresses key machine-learning challenges, such as class imbalance and domain shift.

The rest of the paper is organized as follows. Section \ref{sec_RelatedWork} reviews related work.
Sections \ref{sec_methodology_data} and \ref{sec_methodology_model} describe the proposed methodology (dataset, model architecture and training). 
Section \ref{sec_experiments} presents experimental results and discusses performance across different benchmarks. Finally, Section \ref{sec_conclusions} summarizes the main conclusions.

\section{Related Work} \label{sec_RelatedWork}
\paragraph{Facial age estimation}

Facial aging involves gradual changes in facial proportions, feature shape, and skin texture. Early works explicitly modeled these factors. \citet{Lanitis_2002_FGNET} built a statistical model using principal component analysis of facial landmarks and the isolation of the aging axis in PCA space, while \citet{Han_2013_AgeEstimation} introduced biologically inspired features with a decision tree and support-vector regressors.

Later approaches employed deep learning and convolutional neural networks (CNNs) \cite{LeCun2015_deeplearning}, trained end-to-end without the need for landmarks or handcrafted features \cite{Levi_2015_AgeCNN, Rothe_2018_IMDBWIKI}.
Model sizes range from a few million parameters \cite{Levi_2015_AgeCNN} to over 100M parameters in the case of the VGG-16 architecture used by \citet{Rothe_2018_IMDBWIKI}.
SSR-Net \cite{Yang_2018_SSRNet} is a CNN-based age-regression model that uses a stage-wise strategy, reducing the size (less than 100k parameters) and computational effort significantly.
\citet{Li_2022_CVPR} developed an adaptive label distribution learning, exploiting the ordinal nature of age classes.

Recently, \citet{Paplham_CVPR2024_benchmark} compared state-of-the-art age estimation algorithms and provided an age estimation benchmark, including train/validation/test splits for several facial age estimation datasets.
Considering the mean absolute error averaged over the test set of the respective split, they found that the key to good age-estimation accuracy is a large pre-trained backbone rather than the details of the loss function.
The proposed model of \citet{Paplham_CVPR2024_benchmark} combines the FaRL image encoder \cite{Zheng_2022_FaRL} with a small multilayer perceptron (MLP) for age estimation.
FaRL is a general facial representation model based on the general-purpose CLIP model \cite{Radford_2021_CLIP}, pre-trained with contrastive loss to learn the high-level semantic meaning from image-text pairs for a general facial representation.

\paragraph{Facial underage detection}
Meanwhile, other works have focused on the binary task of distinguishing minors from adults.
\citet{CastrillonSantana_2018} composed a dataset of unrestricted images from a collection of datasets and compared local descriptors and CNNs, i.e., features learned  end-to-end, for underage detection. They found that CNNs slightly outperformed local descriptors, but that score fusion among the two approaches could lead to even better accuracy.
\citet{Anda_2020_DeepUAge} proposed the DeepUAge age estimation model, trained with the VisAGe dataset for accurate age estimation of underage subjects, and
\citet{Chavez_VISAPP2020} studied occluded-eye cases relevant to CSEM.
%
More recently, \citet{Gangwar_2021_AttMCNN} employed CNNs with an attention mechanism and metric learning for pornographic content detection and age-group classification, and combined these into a CSEM detection pipeline.

\section{Overall Underage Benchmark and ASWIFT-20K wild test}\label{sec_methodology_data}

This section introduces the proposed dataset and benchmarks: the Overall Underage Benchmark (Section \ref{ssec:overall-train}), together with the ASORES-39k test and the ASWIFT-20K test, designed to assess model robustness to distribution shift (Section \ref{ssec:eval-protocols}).

\subsection{Overall Underage Benchmark}\label{ssec:overall-train}
\begin{figure}[h]
  \newcommand{\sampleimages}[1]{\subfigure[#1]{%
    \includegraphics[width=0.32\linewidth, trim=0 20 0 20]{Figure1_#1_224x224_samples.jpg}}
  }
  \sampleimages{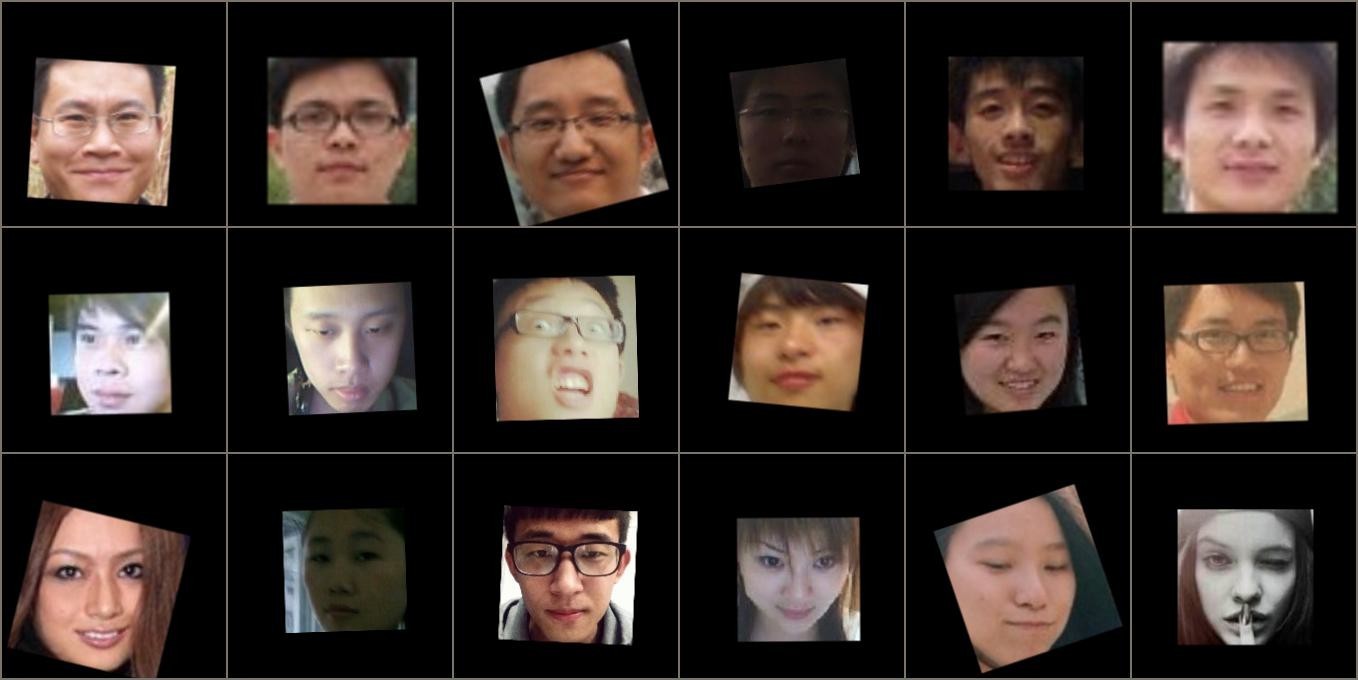} \hfill \sampleimages{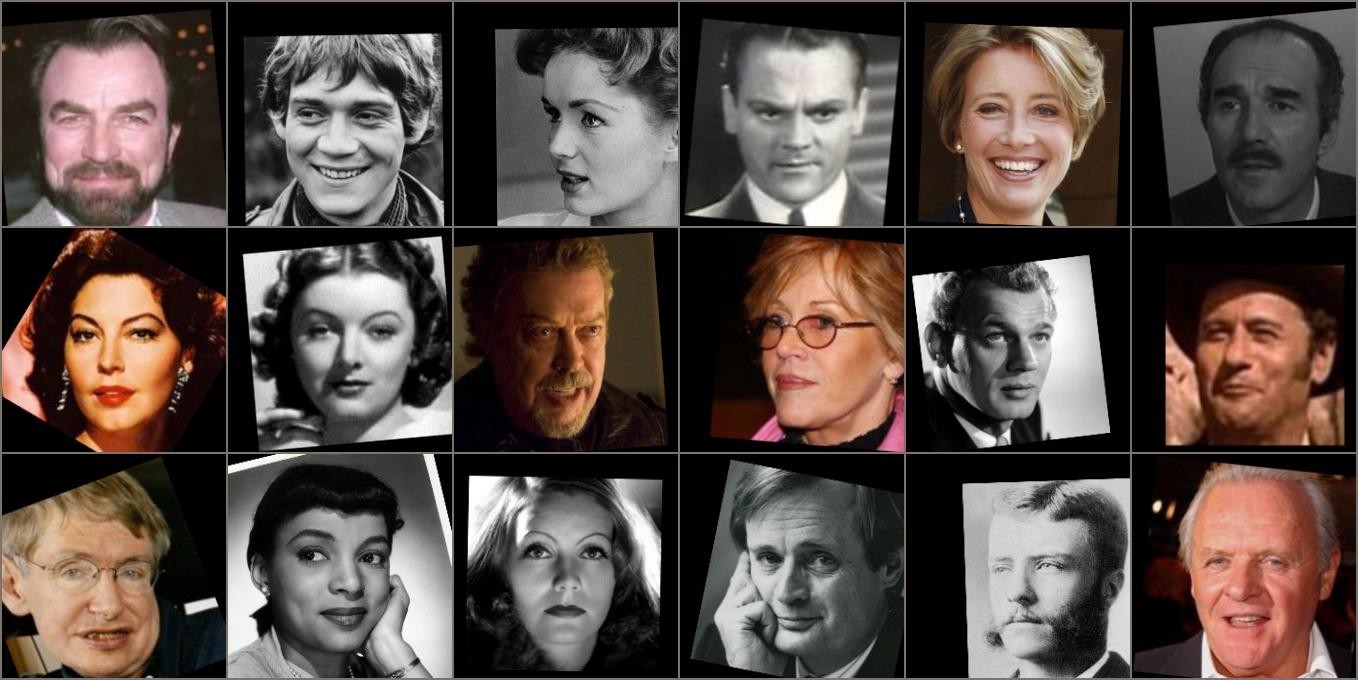} \hfill \sampleimages{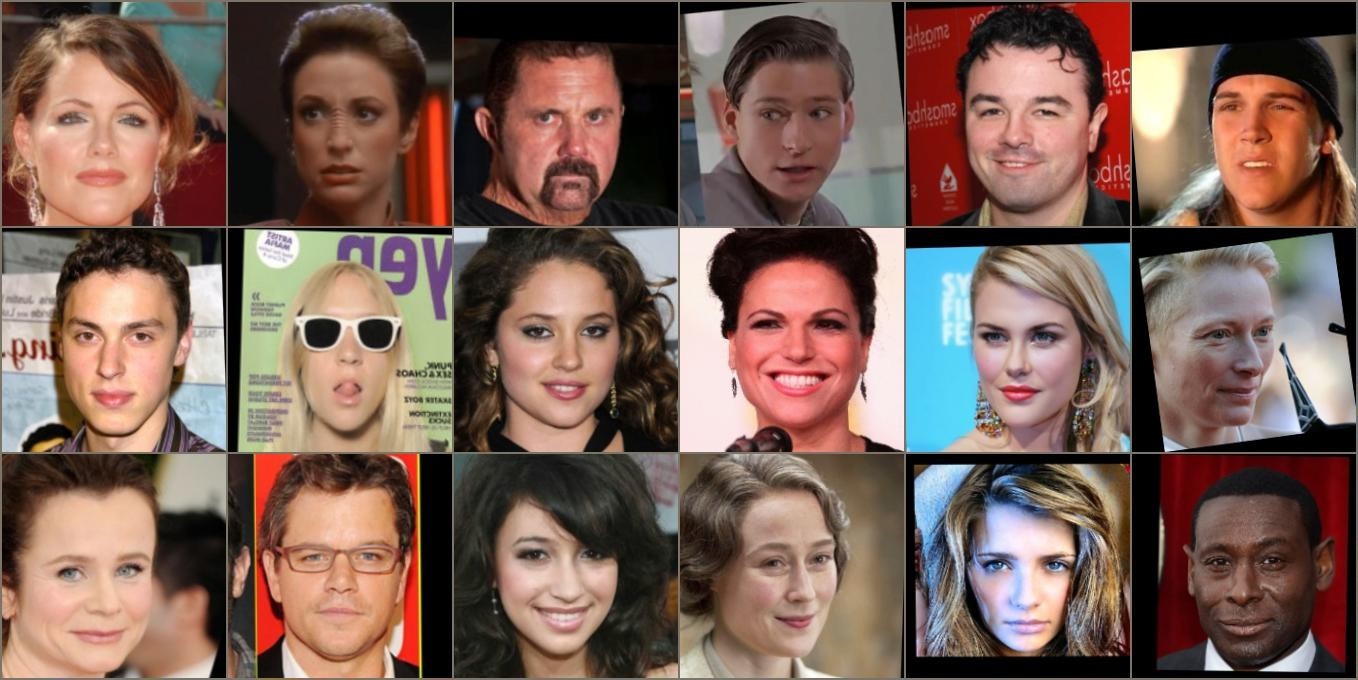} \\
  \sampleimages{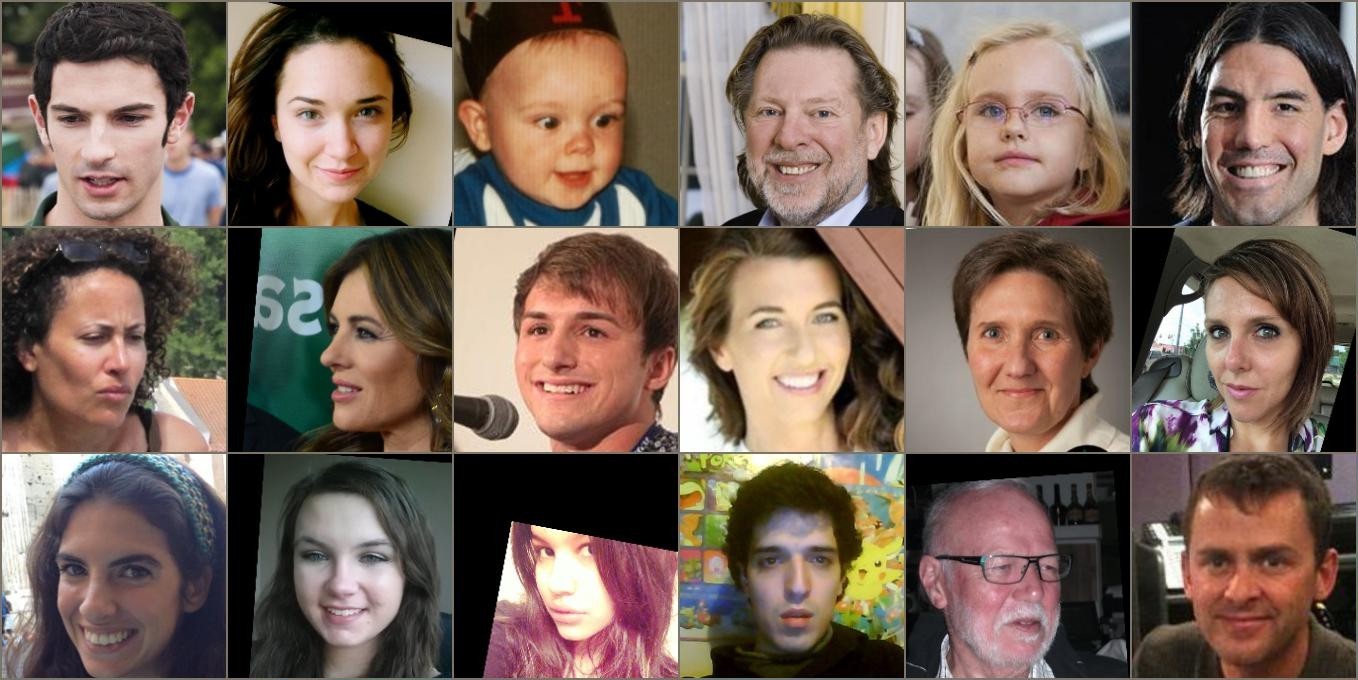} \hfill \sampleimages{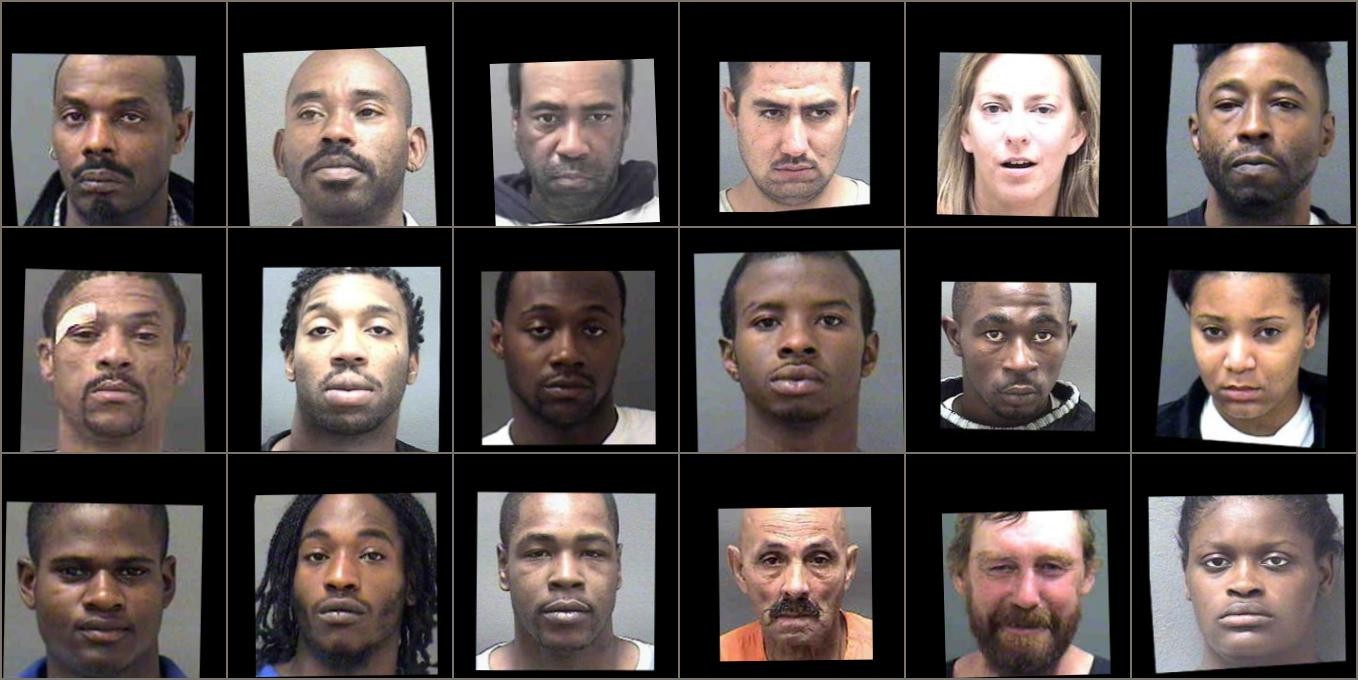} \hfill \sampleimages{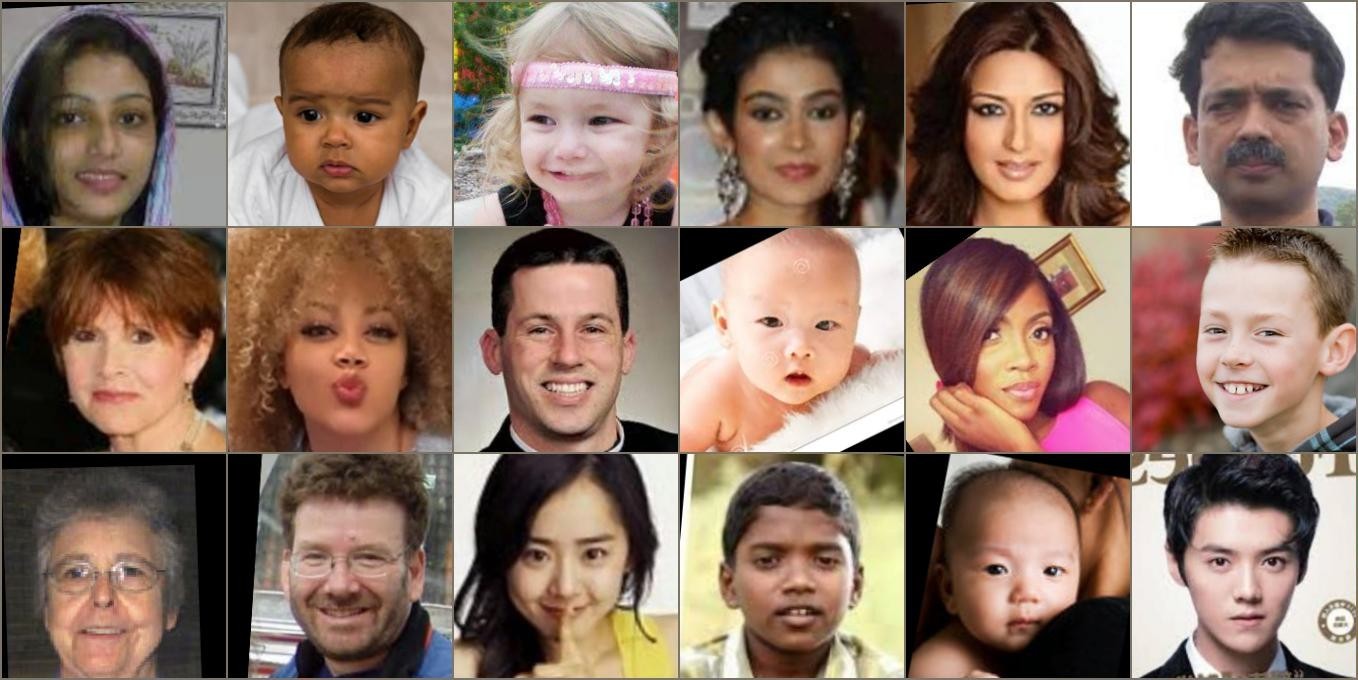}\\
  \caption{Sample face patches from the different age estimation benchmarks, as processed by the model during training, i.e., with the random flipping, scaling, and cropping described in Section \ref{sec_methodology_model}.}\label{fig_dataset_samples}
\end{figure}

For real-world applications of facial age estimation and underage detection, the age algorithm must be reliable across diverse data and generalize well; robustness to distribution shifts (e.g., imaging conditions, subject demographics, poses, facial expressions) is crucial.
Therefore, we merge several benchmarks across the train/validation/test splits provided by \citet{Paplham_CVPR2024_benchmark} into an \emph{overall} benchmark.
The joint training set consists of data from the publicly available datasets
AFAD \cite{Niu_2016_AFAD},
AgeDB \cite{Moschoglou_2017_AgeDB},
CLAP2016 \cite{Agustsson_2017_AppaReal},
the Cross-Age Celebrity Dataset (CACD2000) \cite{Chen_2014_CACD2000},
MORPH Album~2 \cite{Ricanek_2006_MORPH}, and
UTKFace \cite{Zhang_2017_UTKFace}.
These datasets cover variations in age, image quality, pose, and facial expression. Figure \ref{fig_dataset_samples} shows some representative examples.
AFAD contains Asian subjects with low resolution, and the face crops are tight, which leads to padding in the face patches.
MORPH Album~2 contains mugshot images with mostly frontal poses, neutral to negative facial expressions and controlled illumination.
AgeDB, CACD2000, CLAP2016, and UTKFace are unconstrained datasets with diversity in pose, illumination, and expression.
AgeDB also contains monochrome photographs.
Among them, the CACD2000 dataset \cite{Chen_2014_CACD2000} stands out in the sense that it has
noisy age annotations and was not initially intended for age estimation. Therefore, we analyze two benchmark variants, with and without the CACD2000 dataset.

\begin{figure}[ht]
  \centering
  \newcommand{\distrfigure}[1]{\includegraphics[height=3.1cm, trim=2 30 2 20]{#1}}
  \subfigure[\label{fig-age-distribution-train-0}]{
    \distrfigure{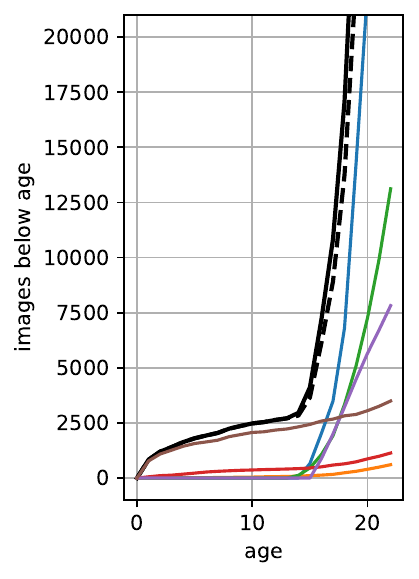}
  }
  \hfill
  \subfigure[\label{fig-age-distribution-train-1}]{
    \distrfigure{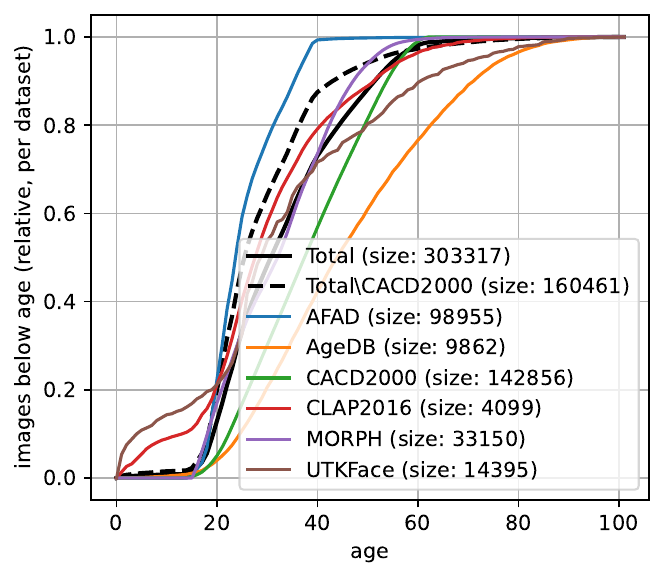}
  }
  \hfill
  \subfigure[\label{fig-age-distribution-test}]{
    \distrfigure{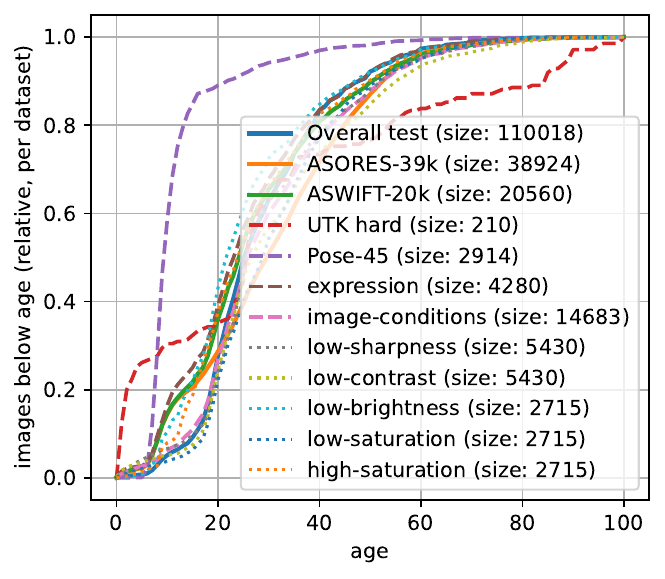}
  }
  \caption{The age distribution of the training and test data. (a) Number of images for each source dataset (colors) and for the merged corpus (black). (b) same curves for the whole age range normalized by the respective dataset sizes.
  (c) Cumulative histogram of the overall test, the ASORES-39k restricted test, and the ASWIFT-20k test and its subsets. Note that the ``Pose-45'' subset, i.e., those faces with a pose angle beyond 45\textdegree{}, comes mostly from the Dartmouth Dataset of Children's Faces and thus contains mostly children.}\label{fig-age-distribution}
\end{figure}%

Figure \ref{fig-age-distribution} shows the age distribution of the overall training set, resolved by source datasets.
Images of subjects below 15 years are only weakly represented, and come mostly from the datasets UTKFace, CLAP2016, and AgeDB.
CACD2000 and AFAD, which are concentrated in the age ranges 15 – 35 and 20 – 45, respectively, show a steep rise beyond 15 years.
The resulting aggregate displays the highly skewed age distribution that motivates the age-balanced resampling scheme, below in Section \ref{sec_meth_alpha_age_equilibration}.

We employ the data splits provided by \citet{Paplham_CVPR2024_benchmark} and use the respective holdout parts of the training data for testing.
Further, we use the FG-NET dataset \cite{Lanitis_2002_FGNET} for testing only, as well as the Dartmouth Database of Children's Faces \cite{Dalrymple_2013_Dartmouth} and the CASIA-Face-Africa dataset collected by the Chinese Academy of Science's Institute of Automation (CASIA) \cite{Muhammad_2021_Casia}.

We used a preliminary model for detecting inconsistencies in the age labels.
We selected the images with the largest training and validation errors\footnote{%
Criteria: (i) worst over-18 classifications.
labeled age $\geq18$: lowest $0.5\%$ of over-18 scores ($1738$ images);
(ii) worst under-18 classifications.
labeled age $< 18$: highest $2\%$ of over-18 scores ($293$ images);
(iii) worst age underestimations.
estimation error $\leq-16$ and estimate $<24$ ($1006$ images);
(iv) worst age over-estimations.
estimation error $\geq9$) and labeled age $<18$ ($990$ images).}
for manual inspection and classified them into the categories \emph{bad age}, \emph{doubtful}, \emph{fine} and \emph{bad image}.
To avoid bias toward our algorithm, doubtful cases were preserved, and only images with clearly wrong age labels and invalid images (non-faces or cartoons) were removed.
In another iteration, we also cleaned the test set (with slightly more tolerant settings). 
In total, 2777 out of 303075  training images, 238 out of 60220 from validation images, and 114 out of 109089 test images were removed.

\subsection{ASWIFT-20k, ASORES-39k and Evaluation Protocols}\label{ssec:eval-protocols}
  \subsubsection{Restricted Overall Test}\label{sec_overall_test}
The combined test bed of Section \ref{ssec:overall-train}
includes the test sets of all age datasets used in this study (AFAD, AgeDB, CACD200, CLAP2016, MORPH Album~2, UTKFace, FG-NET, the Dartmouth database of children's faces, CASIA-Face-Africa).
As described at the end of Section \ref{ssec:overall-train}, the data has been cleaned from the worst label noise.
We use the test split from \citet{Paplham_CVPR2024_benchmark} (fold 0), except for FG-NET, Dartmouth, and CASIA-Face-Africa, which are not used in the training and are used entirely for testing.
The overall test consists of 110k images ($30\%$ from AFAD and $34\%$ from CASIA-Face-Africa).

\paragraph{ASORES-39k, the ``Age Estimation Overall Restricted Test''}
AFAD and CASIA-Face-Africa dominate the overall test.
This is problematic because both present challenging conditions. AFAD has faces cropped tightly and at low resolution, and its age labels seem noisy. At the same time, CASIA-Face-Africa has challenging imaging conditions, including a low-quality webcam and an IR camera. Therefore, we define a \textit{Restricted Overall Test} that excludes these two datasets, which will be considered for the "wild" test case below.
The restricted overall test consists of $39k$ images (38924), distributed across datasets as follows:
AgeDB: $9\%$, CACD2000: $27\%$, CLAP2016: $5\%$, Dartmouth: $16\%$, FG-NET: $3\%$, MORPH: $28\%$, UTKFace: $12\%$.

\subsubsection{ASWIFT-20k, the unconstrained ``Age Estimation Wild-Shifts Test''}\label{sec_wild_test}
Apart from overall high age regression and underage detection performance, it is vital that the accuracy does not break down under real-world conditions, i.e., if variations in imaging conditions, poses, or facial expressions come into play. Therefore, we propose ASWIFT-20k, a \emph{wild} test, where we deliberately select images that deviate from standard conditions.
%
We inspected the $4.8k$ images of the UTKFace test set and manually selected $210$ images that showed interesting poses, facial expressions, and imaging conditions. Since manual selection is tedious and subjective, we turn to automatic selection criteria by pose angles facial expression and image properties. 
The selection criteria are described in detail in the following subsection \ref{sec_selection}.

\begin{table}
    \resizebox{\linewidth}{!}{\begin{tabular}{lrrrrrrrrrr}
\toprule
Criterion & AgeDB & CACD & CLAP & DDCF & FG-NET & MORPH & UTKFace & AFAD & CFA & Total \\
\midrule
UTK hard & 0 & 0 & 0 & 0 & 0 & 0 & 210 & 0 & 0 & 210 \\
Pose-45 & 45 & 68 & 86 & 2521 & 2 & 0 & 16 & 85 & 91 & 2914 \\
expression & 194 & 700 & 156 & 861 & 46 & 165 & 450 & 918 & 790 & 4280 \\
low-sharpness & 356 & 406 & 322 & 3 & 296 & 19 & 1712 & 1326 & 990 & 5430 \\
low-contrast & 393 & 926 & 295 & 0 & 254 & 600 & 646 & 1326 & 990 & 5430 \\
low-brightness & 136 & 262 & 167 & 518 & 65 & 213 & 196 & 663 & 495 & 2715 \\
low-saturation & 945 & 7 & 296 & 0 & 179 & 0 & 130 & 663 & 495 & 2715 \\
high-saturation & 106 & 478 & 231 & 394 & 59 & 3 & 286 & 663 & 495 & 2715 \\
\midrule
Union & 1724 & 2364 & 1058 & 3391 & 590 & 856 & 2687 & 4515 & 3375 & 20560 \\
\bottomrule
\end{tabular}
}
    \caption{Composition of the image-parameters part of the ASWIFT-20k wild test by source datasets. CACD = CACD2000, CLAP = CLAP2016, DDCF = Dartmouth Database of Children's Faces, CFA = CASIA-Face-Africa.}\label{wild-test-composition}
\end{table}

\begin{figure}[hbt]
\includegraphics[width=6cm,trim=0 25 0 10]{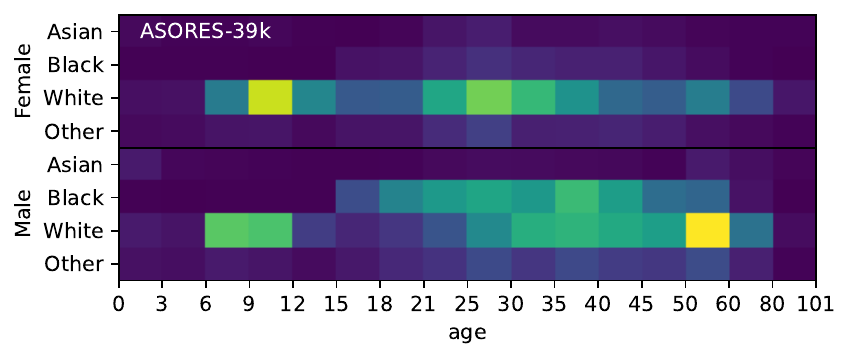}\hfill
\includegraphics[width=6cm,trim=0 25 0 10]{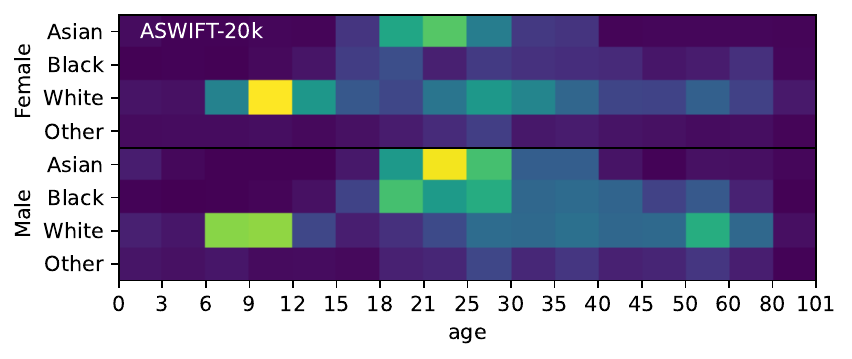}
\caption{Distribution of ASORES-39k and ASWIFT-20k over gender, ethnicity (FairFace \cite{Karkkainen_2021_FairFace}) and age.}\label{fig_aswift_distribution}
\end{figure}

The resulting composition of the ASWIFT-20k wild test in terms of source datasets and selection criteria is summarized in Table \ref{wild-test-composition}.
Figure \ref{fig-age-distribution-test} shows the age distributions of the overall test, the ASORES-39k test and the ASWIFT-20k test and its subsets.
Finally, Figure \ref{fig_aswift_distribution} shows the demographic distributions of the composed test sets ASORES-39k and ASWIFT-20k as heat maps.
While ASWIFT-20k has is wider distribution in the ethnicity dimension than ASORES-39k, both show strong signatures of their respective constituents: blacks are males aged 18 to 60 years (from MORPH), white children of ages 6 to 9 (from Dartmouth), and Asians aged 18 to 30 years (from AFAD).
These imbalances have to be kept in mind when analyzing performance measures on these test sets.

\subsubsection{Selection criteria implemented}\label{sec_selection}
To automate the image retrieval related to \textit{pose angle}, we use the InsightFace Face Analysis\footnote{%
\url{https://github.com/deepinsight/insightface/tree/master/python-package}}
to obtain automatic estimations of the pose angles pitch, yaw, and roll (the roll angle is close to zero because we work with aligned face images).
We use the Euclidean norm of the pose-angle vector as the criterion for the wild test and select 2914 images with a pose angle greater than 45\textdegree.

Furthermore, we separate our test cases by \textit{imaging conditions}.
For all images of the overall test set, we have computed measures for brightness, contrast (mean and standard deviation of the grayscaled face patch, respectively), saturation (mean saturation in HSV color space) and sharpness (variance of Laplacian) of a tight face crop. 
For ASWIFT-20k, we select from the restricted overall test the lowest $8\%$ in \textit{contrast and sharpness}, the lowest $4\%$ in \textit{brightness and saturation}, and the highest $4\%$ in \textit{saturation}.
We have added extreme imaging conditions from the AFAD test set and the CASIA-Face-Africa dataset (only webcam images, not the IR ones) separately and with half the percentages mentioned above (otherwise, these would dominate the wild test too much).
In total, $14683$ images were selected based on the imaging conditions (note that there is some overlap across the different criteria; e.g., many images are both low-contrast and low-sharpness, or both low-contrast and low-brightness).
Table~\ref{wild-test-composition} shows the breakdown into source datasets and selection criteria.

\begin{figure}[tb]
\centering
\subfigure[Selected faces in arousal-valence space]{
  \includegraphics[width=0.62\linewidth,trim=0 20 0 30]{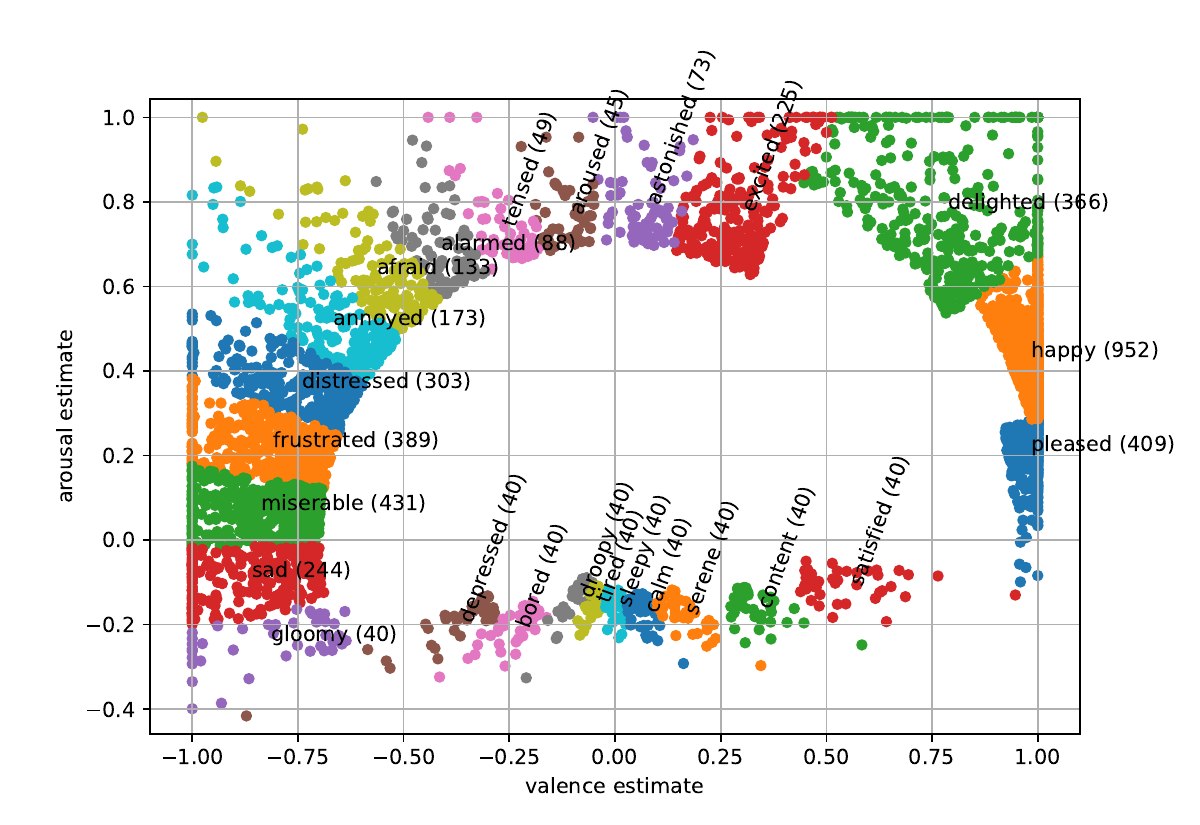}}
\subfigure[Examples]{
  \newcommand{\example}[1]{
    \includegraphics[width=1.8cm]{Figure4b_expression_#1_examples.png}~\raisebox{1ex}{\small #1}}
  \raisebox{2cm}{
  \begin{tabular}{l}
	\example{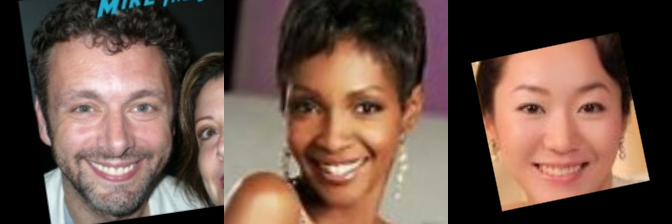} \\
	\example{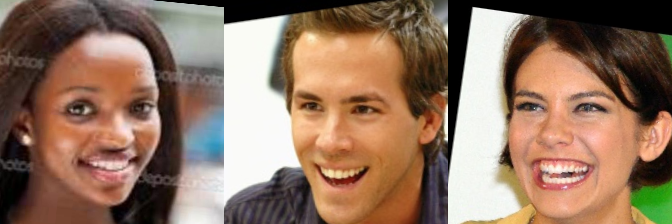} \\
	\example{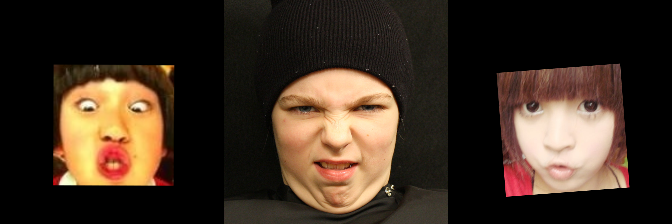} \\
	\example{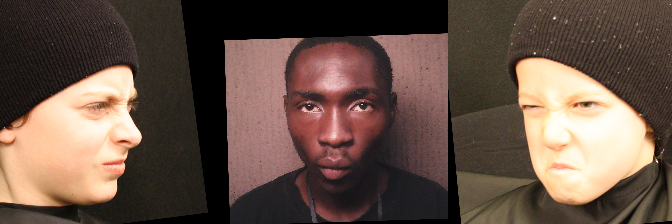} \\
	\example{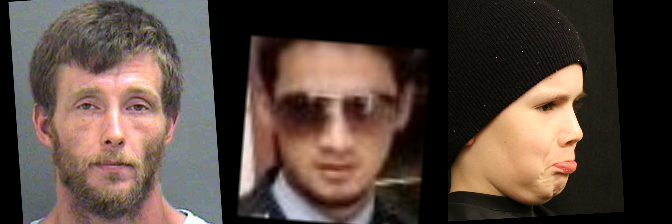} \\
	\example{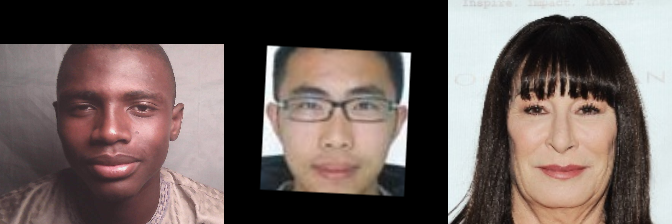} \\
  \end{tabular}}
}
    \caption{Selection of facial expressions for ASWIFT-20k. (a) Distribution of the selected ``extreme'' expressions in arousal-valence space. (b) Example images for some expressions.}\label{fig_extreme_expressions}
\end{figure}

To work with the last criterion, \textit{facial expression},
we use the MorphSet facial-expression estimator by Vonikakis et al.\ \cite{
Vonikakis_2021_expression}, to single out extreme facial expressions.
The MorphSet model operates on frontalized landmarks obtained with the \texttt{dlib} library and provides expression estimates in the valence-arousal space via a partial least-squares regression.
The valence axis ranges from negative/unpleasant to positive/pleasant, and the arousal axis from calm/passive to aroused/active, with neutral in the center.
Figure \ref{fig_extreme_expressions} gives a visual representation of the distribution of the selected face images in the arousal-valence space, together with the number of images chosen for each expression category.
The Euclidean distance from the center is the intensity of the expression, and it is used as a criterion to select extreme expressions.
We select all images with an intensity value above $0.7$ for most expressions in the ``wild'' test. For the expressions \emph{happy}, \emph{pleased}, and \emph{delighted}, which are over-represented in the data, we apply higher thresholds.
For the less common expressions \emph{gloomy}, \emph{depressed}, \emph{bored}, \emph{droopy}, \emph{tired}, \emph{sleepy}, \emph{calm}, \emph{serene}, \emph{content}, and \emph{satisfied} (all with negative arousal), we lower the thresholds such that 40 images are selected.

\section{Underage detection architecture with MultiAge network}\label{sec_methodology_model}
In this work, we pick up the thread of the ``FaRL + MLP'' model introduced by \citet{Paplham_CVPR2024_benchmark} and adapt it to the downstream task of underage detection under real-world conditions.
With 87M parameters, the image encoder of FaRL \cite{Zheng_2022_FaRL}, a 12-layer vision transformer \cite{Dosovitskiy_2021_ViT}, is large compared to early CNNs and ``slim'' models like SSR-Net \cite{Yang_2018_SSRNet}, but still smaller than the VGG-16 of \citet{Rothe_2018_IMDBWIKI}.

\subsection{MultiAge network}\label{sec_MultiAge}

\begin{figure}[tb]
\centering
\includegraphics[width=\linewidth, trim=40 25 0 10]{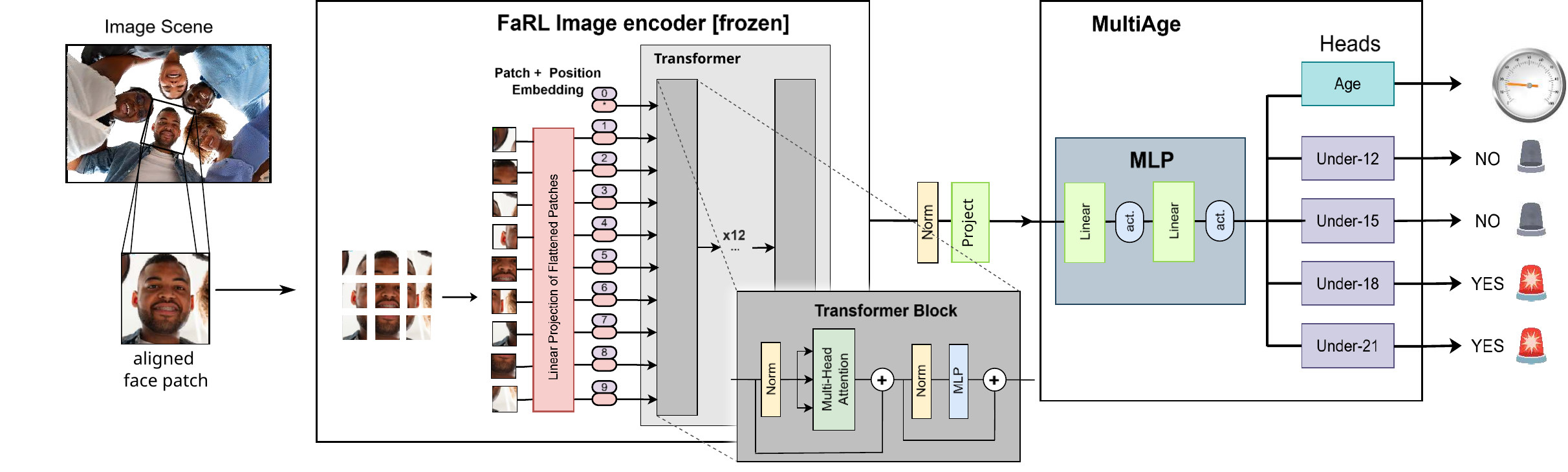}
\caption{Scheme of the model architecture: The face patch extracted from the image scene is divided into subpatches ($14\times14$ non-overlapping patches of $16\times16$ pixels), transformed with the FaRL image encoder \cite{Zheng_2022_FaRL} and then processed by our ``MultiAge'' network. Apart from an age estimate (with its typical estimation error), several outputs indicate whether the subject is likely to be under a given age threshold. As discussed in Section \ref{sec_experiments} below, these underage outputs are more reliable than using the age estimate directly.}\label{fig_architecture}
\end{figure}

Figure \ref{fig_architecture} depicts a graphical summary of the underage detection architecture proposed.
An aligned input \emph{face patch} ($224 \times 224$ pixels) is extracted from an \textit{image scene}.
During training, this is done with random augmentations, randomized crop, and horizontal flip.
The face patch is tokenized into patches of size $16\times16$, linearly projected and fed to the frozen FaRL image encoder~\cite{Zheng_2022_FaRL}, a \mbox{ViT-B/16} transformer with $L\,{=}\,12$ blocks and hidden width $768$~\cite{Dosovitskiy_2021_ViT}.
The normalized and projected output of the FaRL image encoder is denoted $\mathbf{z}\!\in\!\mathbb{R}^{d}$, $d=512$, and constitutes the visual representation passed to the MultiAge network, a light MLP with $k = 5$ task heads, one regressor and four binary detectors, which will be the only trainable part.

The code proposed by \citet{Paplham_CVPR2024_benchmark}, however, shows a deviation from this intended behavior: the dense layers are identical and share the same weights\footnote{Details:  \url{https://github.com/paplhjak/Facial-Age-Estimation-Benchmark/issues/27}}, which limits the expressivity of the network and has a regularizing effect.
We investigate this weight–sharing found in
\citet{Paplham_CVPR2024_benchmark}, expressed as \textit{Weight–sharing MLP} (WS)  against an independent two-layer design with different width configurations, we denoted as \textit{Independent MLP (IND)}.
In the work of \citet{Paplham_CVPR2024_benchmark}, they implemented a weight–sharing MLP as follows:
    \begin{equation}
    \begin{aligned}
        \mathbf{h}_{\mathrm{WS}} &= ReLU\!\bigl(\mathbf{W}\,\mathbf{z} + \mathbf{b}_1\bigr), &
        \mathbf{o}_{\mathrm{WS}} &= ReLU(\mathbf{W}\,\mathbf{h}_{\mathrm{WS}} + \mathbf{b}_2),\\
    \end{aligned}\label{eq:Paplham}
    \end{equation}
with a single $\mathbf{W}\!\in\!\mathbb{R}^{d\times d}$ used twice. This reduces the parameter count from $2d^2$ to $d^2$ and regularizes the mapping, but constrains its rank to $d$.

However, we hypothesize that the MultiAge network with the Independent MLP we propose, denoted as:
\begin{equation}
\begin{aligned}
    \mathbf{h}_{\mathrm{IND}} &= ReLU\!\bigl(\mathbf{W}_1 \, \mathbf{z} + \mathbf{b}_1\bigr),
     \mathbf{W}_1\!\in\!\mathbb{R}^{m\times d}, &
    \mathbf{o}_{\mathrm{IND}} &= ReLU(\mathbf{W}_2 \,\mathbf{h}_{\mathrm{IND}} + \mathbf{b}_2),
     \mathbf{W}_2\!\in\!\mathbb{R}^{d\times m} \\
 \end{aligned}\label{eq:Gaul_et_al}
\end{equation}
separates $\mathbf{W}_1$ and $\mathbf{W}_2$, increasing the rank constraint and allowing the model to learn task-specific projections, by letting the network learn more parameters.

In the baseline model \cite{Paplham_CVPR2024_benchmark}, $\mathbf{o}$ is fed into the age estimation head, implemented as classification in $c = 102$ age classes for ages 0 to 101, trained with cross-entropy loss \cite{Paplham_CVPR2024_benchmark}, and the predicted age $\hat{y}$ is given as
\begin{equation}
	\hat{y}=\sum_{a=0}^{101} a p^\mathrm{age}_a,
	\qquad p^\mathrm{age} = \mathrm{softmax}\left( \mathbf{w}_{\text{age}}\mathbf{o}+b_{\text{age}}\right),
	\qquad \mathbf{w}_{\text{age}}\in\mathbb{R}^{c\times d}.
	\label{eq:age_regression}
\end{equation}
To this point, the model estimates the age $\hat{y}$, which can be used for underage detection, though it is limited by estimation errors.

In the spirit of multitask learning \cite{Caruana_1997_Multitask}, we combine \emph{age estimation} task with binary discrimination tasks of underage detection with the binary discriminators for over/underage discrimination.
On the one hand, these underage discriminators are of immediate use in applications (Fig.\ \ref{fig_architecture}), on the other hand, this leverages synergies between the age estimation task and the underage detection tasks, and the binary discrimination should help to guide the focus of the model to the age range of interest.

Thus, for both cases of MLP architectures, $\mathbf{o}$ of Eq.\ \ref{eq:Paplham} or Eq.\ \ref{eq:Gaul_et_al} is forwarded to the age estimation head (Eq.\ \ref{eq:age_regression}) and to $k{-}1$ sigmoid-activated binary heads with thresholds $A=\{12,15,18,21\}$.
The binary heads 
are given as:
\begin{equation}
    \begin{aligned}
        u_{k} = \mathbf{w}_{k}^{\top}\mathbf{o}+b_{k},\quad & \mathbf{w}_{k}\!\in\!\mathbb{R}^{d},\quad &
        p_{k} = \sigma(u_{k}),\quad & k=1,2,3,4,
    \end{aligned}
    \label{eq:binary_underage_heads}
\end{equation}
where $u_{k}$ and $p_{k}$ are the intermediate activation and final output of each head, respectively, and \(\sigma(\cdot)\) is the sigmoid.

Only the MultiAge network with Independent MLP and its heads are trainable (less than a million parameters); the 87 M-parameter FaRL backbone remains frozen.

\subsection{$\alpha$-reweighted focal loss and age-balanced resampling}\label{sec_meth_alpha_age_equilibration}
Minors are underrepresented in the training set, which tends to bias the trained model towards adults. This is detrimental to our objective of faithfully detecting minors and estimating their age.
To mitigate this imbalance, we propose a variation of the focal loss of Lin et al.\ \cite{Lin_ICCV_2017_FocalLoss}, an $\alpha$-reweighted focal loss variant to prioritize minority class examples.
In the case of several underage detectors with different age levels, they exhibit different class imbalances and different values of the correction parameter $\alpha$.

We formally define the proposed $\alpha$-reweighted focal loss as:
\begin{equation}
\mathcal{L}_{\alpha}(y, \hat{y}) = -\alpha_{y}(1 - \hat{y})^\gamma y\log(\hat{y}) - (1 - \alpha_{y})\hat{y}^\gamma (1 - y)\log(1 - \hat{y}),
\label{eq:alpha_FocalLoss}
\end{equation}%
where $\hat{y}$ is the predicted probability, $y$ is the binary ground-truth, and $\gamma$ is the focusing parameter (set to $\gamma=2$ following~\cite{Lin_ICCV_2017_FocalLoss}, for $\alpha = 1$, $\gamma=0$, Eq.\ \ref{eq:alpha_FocalLoss} reduces to the binary cross-entropy loss). The weighting term $\alpha_y$ we propose is computed as follows:

\begin{equation}
\alpha_{y} =
\begin{cases}
\sqrt{{N_{\text{majority}}}/{N_{\text{minority}}}}, & \text{if } y=1\text{ (minority)}, \\[6pt]
\sqrt{{N_{\text{minority}}}/{N_{\text{majority}}}}, & \text{otherwise.}
\end{cases}\label{eq:alpha_parameter}
\end{equation}

\paragraph{Age-balanced resampling}
\citet{An_2021_Resampling} have shown that, for training methods like stochastic gradient descent, reweighting the training losses can have detrimental effects.
Thus, we additionally propose an age-balanced mini-batch resampling scheme.  We employ an age-equalized training scheme in which the age label determines the statistical weight for each image to be selected for the current minibatch during training. The age range is partitioned into $n$ age groups, specifically $n=12$, with age boundaries at 4, 8, 12, 16, 20, 24, 28, 32, 36, 42, and 50. Then, the weight factors for each age group are chosen inversely proportional to the number of images in the age group.
%
Formally, the probability of sampling from age-bin $i$ is set as:
\begin{equation}
P(i) = \frac{1/n_i}{\sum_{j=1}^{n}(1/n_j)},
\label{eq:sampling_age_bin}
\end{equation}
where $n_i$ denotes the number of samples in age-bin $i$. This strategy ensures each age bin contributes equally during training, effectively stabilizing gradient updates and promoting better generalization for underrepresented age groups.

\subsection{Mitigating the impact of edge cases}\label{sec_age_gap}
Considering the binary discrimination task of underage detection for a given age threshold as a function of age, it is clear that misclassifications are most likely to happen for subjects close to the age threshold.
For example, the age estimate for a subject of $17$ years and $11$ months may be easily confused with that of a subject of $18$ years and one month. At the same time, these edge cases may produce strong gradients and confuse the model during training.
Therefore, we propose an \emph{age-gap} hyperparameter, defining an interval around the age threshold; patterns with age in the age gap are neither positive nor negative and do not contribute to the loss of the respective discriminator.
Note, however, that due to the multitask setup and the shared representation $\mathbf{o}$, examples from the age gap still contribute indirectly to the training of the respective underage detection task.

Formally, for a given threshold $T$ and gap $G$, the binary ground-truth label is redefined as follows:
\begin{equation}
y_{\text{binary}}(a) =
\begin{cases}
0, & a \geq T + G \quad\text{(adult)}, \\
1, & a < T - G \quad\text{(underage)}, \\
\text{none}, & T - G \leq a < T + G,
\end{cases}
\end{equation}
where $a$ denotes the actual age.
The value of $G$ is a hyperparameter that has to be validated empirically.
Too large age gaps would remove too many informative training examples, while too small gaps may provide insufficient protection against noisy gradients during training. 

The facial appearance changes more rapidly at young ages than at older ages \cite{Geng_2014_ALDL}.
Therefore, it is natural to define the age gap in \emph{relative} terms with a scale parameter $f>1$, excluding a gap from $T/f$ to $T f$ for the under-$T$ detector.
Empirically, we find optimal results with $f=6/5$ (see Section \ref{sec_mlp} below). 
Taking into account the integer nature of the age labels through rounding towards zero, this results in the age gaps
\begin{equation}
12 + [-2, +2],\quad
15 + [-2, +3],\quad
18 + [-3, +3],\quad
21 + [-3, +4]
\label{eq_rel_age_gap}
\end{equation}
for $T=12, 15, 18, 21$, respectively.

\section{Experiments and Discussion} \label{sec_experiments}

This section presents an empirical evaluation. First,  we reproduce a baseline age estimator (Section \ref{sec_experiments_AgeEstimationBase}), then explore training on the unified dataset (Section \ref{sec_experiments_overalltraining}). Section \ref{sec_experiments_under18} investigates under-18 detection under various training settings, while Section \ref{sec_experiments_12_1518} extends the results to other thresholds and explores architectural variations.

\subsection{Age Estimation Baseline}
\label{sec_experiments_AgeEstimationBase}

To establish a performance baseline, we reproduce the state of the art in age estimation by running the ``FaRL + MLP'' model of \cite{Paplham_CVPR2024_benchmark} under several deployment conditions. Unlike the original approach, we omit the pre-training of MLP on the IMDB-WIKI dataset to simplify the training process and isolate the impact of our contributions.

To illustrate the effect of the mismatch between training and deployment conditions, we set up an experiment in which the model is trained on different datasets and applied to other scenarios.
Table \ref{tab_age_benchmark} reports the Mean Absolute Error (MAE) metric, which refers to the absolute value of the deviation between estimated age and true age, averaged over the respective test dataset.

\begin{table} [ht]
\resizebox{\linewidth}{!}{\begin{tabular}{lrrrrrrr}
\toprule
Benchmark &  AgeDB &  AFAD &  CACD2000 &  CLAP2016 &  FG-NET &  MORPH &  UTKFace \\
\midrule
    AgeDB &   5.71 &  6.07 &      7.14 &      9.08 &    7.32 &   4.77 &    10.71 \\
     AFAD &  14.45 &  3.18 &     10.48 &      8.91 &   14.01 &   6.90 &    11.96 \\
 CACD2000 &  14.53 & 14.07 &      4.24 &      9.52 &   23.13 &   9.33 &    11.87 \\
 CLAP2016 &   7.56 &  4.15 &      6.61 &      3.29 &    4.39 &   4.78 &     4.74 \\
    MORPH &  11.27 &  5.19 &     10.71 &      7.10 &    8.36 &   3.16 &     9.36 \\
  UTKFace &   7.05 &  4.94 &      6.91 &      4.05 &    4.66 &   5.02 &     3.87 \\
\bottomrule
\end{tabular}
}
\caption{Mean Absolute Error (MAE) (in years) obtained in the baseline setting of the Facial-Age-Estimation-Benchmark of \cite{Paplham_CVPR2024_benchmark} with a ``FaRL + MLP'' model.
Each row corresponds to a training set (benchmark), and each column to a test dataset (a holdout set in the case of intra-dataset evaluation).
All rows are averaged over a five-fold cross-validation, except CACD2000 and CLAP2016, where only one split is used.}\label{tab_age_benchmark}
\end{table}

Table \ref{tab_age_benchmark_diff} shows the differences with respect to the values reported by \citet{Paplham_CVPR2024_benchmark}. As mentioned above, their experiments involved pre-training the MLP, whereas ours did not.
This is probably the reason why we measure worse accuracy in the rows ``CACD2000'' and ``MORPH'', i.e.\ for models trained with these two datasets, one with label noise and one with limited variability.
In some cases, we measured lower errors (negative differences in Table \ref{tab_age_benchmark_diff}); thus, in general, we reach a similar accuracy as \citet{Paplham_CVPR2024_benchmark}, even without pre-training the MLP.

\begin{table}  [h!]
\resizebox{\linewidth}{!}{\begin{tabular}{lrrrrrrr}
\toprule
{} &  AgeDB &  AFAD &  CACD2000 &  CLAP2016 &  FG-NET &  MORPH &  UTKFace \\
\midrule
AgeDB    &   0.07 & -1.75 &     -0.27 &     -0.24 &   -1.83 &   0.04 &     0.79 \\
AFAD     &  -1.96 &  0.06 &     -0.47 &      0.34 &    1.77 &   0.28 &     0.32 \\
CACD2000 &   3.21 &  4.99 &      0.28 &      0.95 &    3.50 &   2.77 &     0.60 \\
CLAP2016 &   0.06 & -0.19 &      0.04 &     -0.09 &   -0.56 &   0.31 &    -0.11 \\
MORPH    &   2.87 &  0.52 &      3.26 &      0.89 &   -0.92 &   0.12 &     0.43 \\
UTKFace  &  -0.11 &  0.25 &     -0.02 &      0.03 &   -0.41 &   0.26 &    -0.00 \\
\bottomrule
\end{tabular}
}
\caption{Difference of the Mean Absolute Error (MAE) given in Table \ref{tab_age_benchmark} with respect to the values given in \cite[Table 6, line ``FaRL + MLP'']{Paplham_CVPR2024_benchmark}.
For some entries, our accuracy is worse (positive difference); for others, it is better (negative numbers).}\label{tab_age_benchmark_diff}
\end{table}

\subsection{Age estimation with an overall training set}
\label{sec_experiments_overalltraining}

Apart from the individual benchmarks, we also employ the overall benchmark as described in Section \ref{ssec:overall-train}, which is constructed as a joint training set together with consistent holdout sets for validation and testing.
\begin{figure}[tb]
\centering
\includegraphics[width=0.7\linewidth,trim=0 20 0 40]{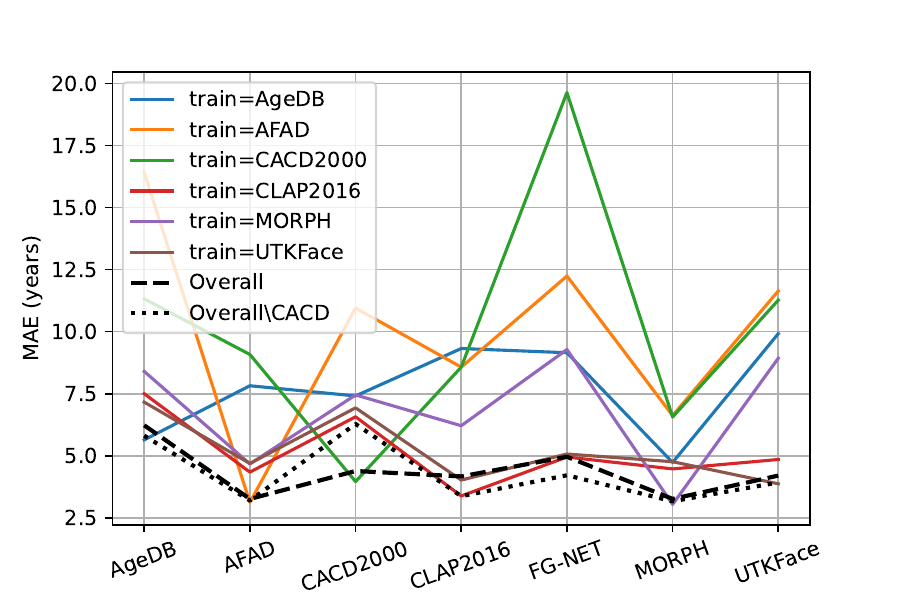}
\caption{Graphical representation of the MAE (Mean Absolute Error) of the model trained with the overall training (black) set compared to those trained for one specific benchmark. The figure compares the results obtained with separated training datasets (Table \ref{tab_age_benchmark}) and those obtained with composed training sets (Table \ref{tab_age_overall}).}\label{fig_benchmark_overall}
\end{figure}
\begin{table} [tb]
  \resizebox{\linewidth}{!}{\begin{tabular}{lrrrrrrrc}
\toprule
   Benchmark &  AgeDB &  AFAD &  CACD2000 &  CLAP2016 &  FG-NET &  MORPH &  UTKFace & \\
\midrule
     Overall &   6.23 &  3.26 &      4.39 &      4.17 &    4.96 &   3.26 &     4.20 & (0) \\
Overall\textbackslash CACD &   5.81 &  3.20 &      6.28 &      3.37 &    4.21 &   3.17 &     3.93 & \\
\bottomrule
\end{tabular}
}
  \caption{Mean Absolute Error (MAE) (in years) obtained with the overall training sets.
  The rows show the overall benchmarks ``Overall'' and ``Overall without CACD'' trained with the union of six and five datasets, respectively, and evaluated on the test set, averaged over five folds. The mark (0) is for future reference.}\label{tab_age_overall}
\end{table}
We consider two variants, the overall training set with and without the CACD data.
The test errors obtained with the different test datasets are listed in Table \ref{tab_age_overall} and represented graphically in Figure \ref{fig_benchmark_overall}.
The model trained on the Overall training set achieves low age-regression errors across all datasets.
Indeed, the model trained on the Overall training set achieves accuracies close to the respective intra-dataset accuracies, indicating that the capacity of the ``FaRL + MLP'' model is sufficient to simultaneously adapt to the data distributions of the individual benchmarks, providing a ``one fits all'' model.

Removing CACD from the training set (Overall$\backslash$CACD) improves MAE by $0.42$ years in AgeDB, $0.80$ years on CLAP2016, and $0.28$ years on FG-NET, confirming that noisy annotations degrade generalization. The regression error increases on the CACD2000 test, becoming an out-of-distribution test without CACD in the training.

\subsection{Under-18 detection}
\label{sec_experiments_under18}

To assess how different training settings and training strategies affect under-18 detection, we evaluate four variants of our pipeline: \textit{data cleaning} (whether the training data was cleaned or raw); \textit{dataset inclusion} (if the noisy CACD2000 dataset was included or not); \textit{sampling strategy} (whether age-balanced resampling was applied in training); and \textit{model architecture} (if a dedicated binary under-18 head was added).


For the binary discrimination task of under-18 detection, under-18 and over-18 are the positive and the negative class, respectively, and the confusion matrix is given by $((\TP, \FN), (\FP, \TN))$, where F/T stands for false/true and P/N for positive/negative.
The most important error measure is the false over-age detection rate, i.e.\ the fraction of underage subjects (positives) that are incorrectly classified as adults (negatives), also known as the false negative rate $\FNR = \FN/(\TP + \FN)$.
This value should be low with high priority.
In detection-error tradeoff (DET) curves, the false adult detection rate $\FNR$ is plotted against the false positive rate $\FPR = \FP/(\TN + \FP)$ or false underage detection rate, parametrized by the threshold applied to the confidence score (or the age estimate used to separate underage from adults).

Recall and precision offer a complementary view on the confusion matrix.
Recall $= \TP / (\TP + \FN) = 1-\FNR$ measures whether all minors were detected correctly. It should be high with high priority.
Precision $= \TP / (\TP + \FP)$ measures whether all detected minors are really minors. It should also be high, but with lower priority.
The $F\beta$ scores interpolate between Recall and Precision. The formula $F\beta = (1 + \beta^2) / (1/\mathrm{precision} + \beta^2/\mathrm{recall})$ reduces to the harmonic mean for F1 and gives higher priority to recall in the case of F2, as it is adequate for the use case of CSEM detection.

\subsubsection{Impact of the training set and training scheme}
We consider the Overall training set of \ref{ssec:overall-train}, with and without label noise cleaning (Section \ref{ssec:overall-train}), as well as a variant that excludes the CACD2000 dataset from the training.
We add a dedicated underage (under-18) detector to the model and train it with binary cross-entropy loss, optionally with the age-balanced resampling (Section \ref{sec_meth_alpha_age_equilibration}).

\begin{figure}[tb]
	\centering
	\includegraphics[width=0.85\linewidth, trim=0 20 0 20]{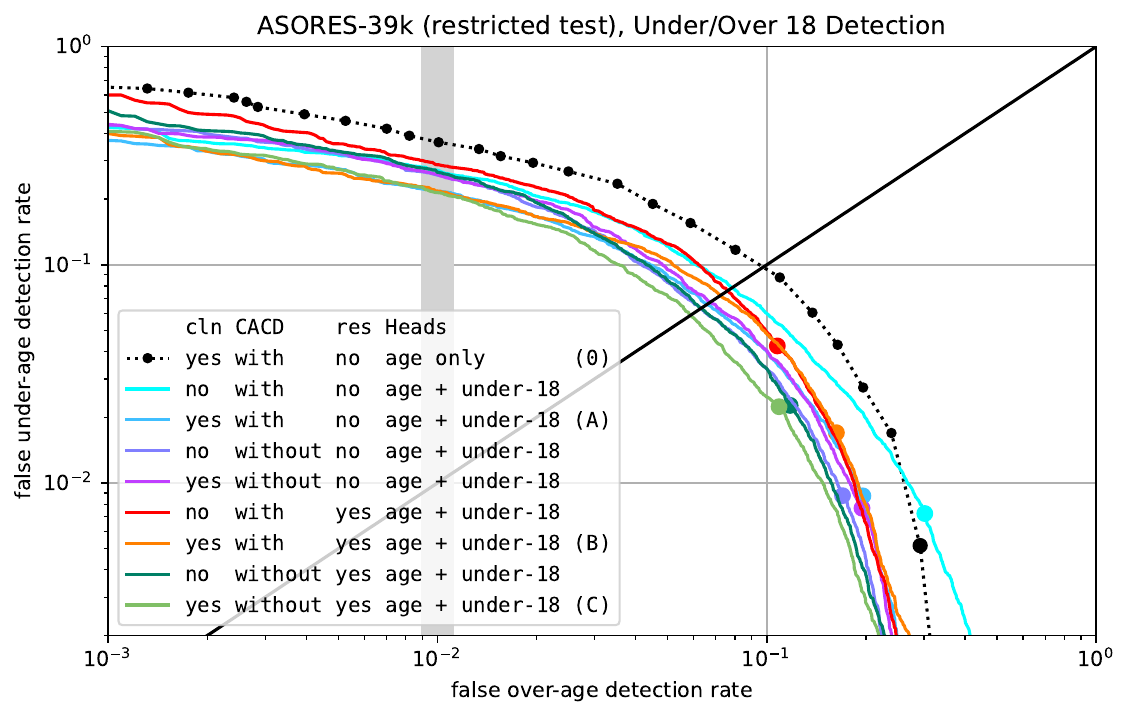}
	\caption{Under/over 18-years discrimination accuracy obtained under different training schemes (cln: cleaned or not; CACD: with or without the CACD part of the training set; res: using resampling according to Eq.\ \ref{eq:sampling_age_bin} or not and with or without a dedicated under/over 18 discriminator).
	Common parameters: $\alpha=1, \gamma=0, G=0$.
    The default working points, marked with thick dots, are ``age estimate = 18'' for the age estimator and over age probability = 0.5 for the binary under-18 discriminator. Shaded area: desired range of operation, false over-age detection rate $\approx 10^{-2}$. The models (0, A, B, C) are marked for future reference.}\label{fig-over-18-training-set-det}
\end{figure}

Figure \ref{fig-over-18-training-set-det} shows the detection error tradeoff (DET) curve for under/over 18-year discrimination measured on the ASORES-39k (restricted) test (Section \ref{sec_overall_test}).
Black dots with dotted lines (``age only'') correspond to the row marked as (0) in Table \ref{tab_age_overall}, where the age estimate is used for under-18 discrimination.
In the other curves, the output of the under-18 head of the network is used, and the curves are parameterized by the threshold applied.
Note that the default working points are strongly biased towards false-over-age detections because the training data contains much more adults than underage subjects (see age distribution in Figure \ref{fig-age-distribution-train-1}, curve ``Total'').
To achieve low false underage detection rates, as required in practice, the threshold must be adjusted significantly.
For example, to achieve a false-adult detection rate as low as $10^{-2}$, the threshold of the plain age estimator has to be raised from 18 to 32 years.

The other curves of Figure \ref{fig-over-18-training-set-det} show how the under-18 discrimination accuracy is improved by (1) adding a binary discriminator as a second head to the model, which is trained for under-18 discrimination (curves with ``+ under-18'' in the legend); (2) modifying the training set by cleaning (end of Section \ref{ssec:overall-train}) and/or by removing the noisy CACD2000 dataset from the training; 
(3)applying age-equalized resampling during training (``res = yes'', described in Sec.\ \ref{sec_meth_alpha_age_equilibration}).
%
Three setups for the training perform equally well at the working point of false-adult detection rate = $10^{-2}$ (the gray region in Figure \ref{fig-over-18-training-set-det}), the FaRL + MLP architecture with age estimation and under-18 heads, trained with the cleaned ``Overall'' training set (A), additionally with resampling (B), and omitting the CACD data in the training (C).
Out of these, the last one, model C, outperforms the others  in a broad range of the DET curve, in particular at the point of equal error rate (diagonal of the plot).

\newcommand{\ModelD}{model D}

\subsubsection{Additional underage detectors}\label{sec_alpha_4discs}
\begin{figure}[htb]
	\centering
	\includegraphics[width=0.85\linewidth,trim=0 20 0 20]{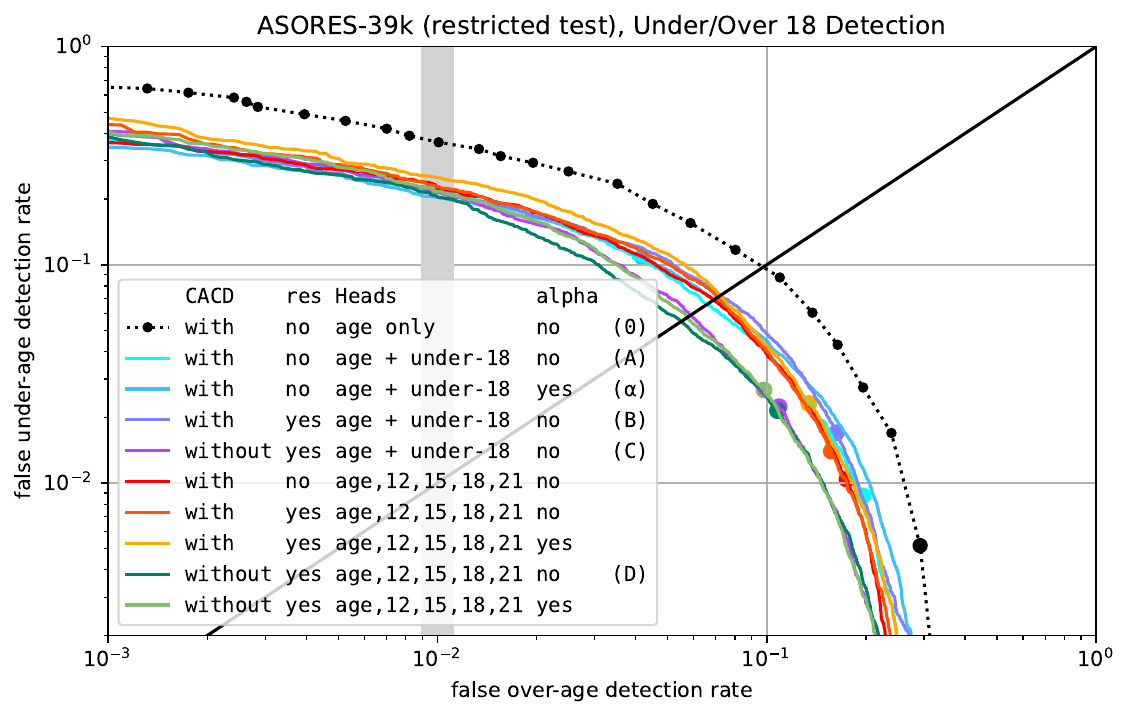}
	\caption{Under/over 18-year discrimination accuracy.
    Variation points:
    use CACD2000 in the training or not;
    resampling (res) according to Eq.\ \ref{eq:sampling_age_bin} or not;
    prediction head setup (Heads);
    reweighting (alpha) according to Eq.\ \ref{eq:alpha_parameter} or $\alpha_{y}=1$.
    Common parameters: cleaned training sets (cln=yes in Figure \ref{fig-over-18-training-set-det}), $\gamma=0, G=0$.}\label{fig-over-18-det-alpha-4discs}
\end{figure}
To improve the under-18 discrimination, we explore the addition of binary detectors by exploring the next two variation points:
(1) the addition of further over-age discriminators (i.e., underage detectors at age levels 12, 15, 18, 21 years) to increase the focus on the underage region and (2) the configuration of the $\alpha$ parameter applied in the loss function to correct for the remaining class imbalance (Section \ref{sec_meth_alpha_age_equilibration}).

Figure \ref{fig-over-18-det-alpha-4discs} shows DET curves comparing models with only an under-18 head to those with additional discriminators on ASORES-39k.
The first four curves, in Figure \ref{fig-over-18-det-alpha-4discs}, are taken from the previous Figure \ref{fig-over-18-training-set-det}.
The curves show the under-18 discrimination accuracy of models that incorporate further underage discriminators at age levels 12, 15, and 18 years.
Training the model with resampling \emph{and} the $\alpha$ parameter (light orange and light green) turns out worse than using just the resampling (dark orange and dark green).
Best accuracy for the under-18 discrimination task is achieved by \emph{\ModelD{}}, i.e.\ using the training set consisting only of AFAD,
AgeDB, CLAP2016, MORPH Album 2, and UTKFace (but not CACD), trained with age-balanced resampling (Sec.\ \ref{sec_meth_alpha_age_equilibration}) and with further underage discriminators (but without further tuning of the $\alpha$ parameter).

For a more extensive comparison of accuracy measures of different age models in different test cases, we fix the threshold of the underage discriminators as the point where the false overage detection rate is at $10^{-2}$, i.e.\ at Recall = 0.99 on ASORES-39k.
In Figures \ref{fig-over-18-training-set-det} and \ref{fig-over-18-det-alpha-4discs}, this is marked by the gray area, and the figure of merit is a low false under-18 detection rate.
In the following, we switch to the accuracy measures Recall and Precision and to their fusion in the F1 and F2 scores.

\begin{table}[htb]
	\renewcommand{\labelA}{\rotatebox{0}{\parbox{1.cm}{\centering age}}}
    \renewcommand{\labelB}{\rotatebox{0}{\parbox{1.cm}{\centering age + u-18}}}
    \renewcommand{\labelC}{\rotatebox{0}{\parbox{1.cm}{\centering age + 12, 15, 18, 21}}}
    \resizebox{\linewidth}{!}{\begin{tabular}{ccccrrrrrrrrc}
\toprule
 &  &  &  & \multicolumn{4}{c}{ASORES-39k (restricted)} & \multicolumn{4}{c}{ASWIFT-20k (unconstrained)} &   \\
Heads & CACD & res & $\alpha$  & Pr & F1 & F2 & Re & Pr & F1 & F2 & Re &  \\
\midrule
\labelA & with & no & no & 0.454 & 0.623 & 0.801 & 0.990 & 0.377 & 0.544 & 0.742 & 0.981 & (0) \\
\midrule
\multirow[c]{4}{*}{\labelB} & with & no & no & 0.583 & 0.734 & 0.869 & 0.990 & \bfseries 0.486 & \bfseries 0.651 & \bfseries 0.817 & 0.984 & (A) \\
 & with & no & yes & \bfseries 0.598 & \bfseries 0.746 & \bfseries 0.875 & 0.990 & \bfseries 0.485 & \bfseries 0.650 & \bfseries 0.817 & \bfseries 0.986 & ($\alpha$) \\
 & with & yes & no & 0.582 & 0.733 & 0.868 & 0.990 & 0.464 & 0.631 & 0.803 & 0.983 & (B) \\
 & without & yes & no & \bfseries 0.587 & \bfseries 0.737 & \bfseries 0.870 & 0.990 & \bfseries 0.474 & 0.639 & 0.807 & 0.980 & (C) \\
\midrule
\multirow[c]{5}{*}{\labelC} & with & no & no & 0.575 & 0.727 & 0.865 & 0.990 & 0.474 & 0.640 & 0.811 & \bfseries 0.986 &  \\
 & with & yes & no & 0.570 & 0.724 & 0.863 & 0.990 & 0.436 & 0.604 & 0.787 & 0.985 &  \\
 & with & yes & yes & 0.547 & 0.704 & 0.852 & 0.990 & 0.428 & 0.597 & 0.782 & 0.985 &  \\
 & without & yes & no & \bfseries 0.597 & \bfseries 0.744 & \bfseries 0.875 & 0.990 & 0.474 & \bfseries 0.641 & \bfseries 0.813 & \bfseries 0.990 & (D) \\
 & without & yes & yes & 0.583 & 0.734 & 0.869 & 0.990 & 0.464 & 0.630 & 0.804 & 0.985 &  \\
\bottomrule
\end{tabular}
}
\caption{Quantitative comparison of under-18 detection accuracy of the models shown in Figure \ref{fig-over-18-det-alpha-4discs}.
The discriminator threshold is fixed such that the false-adult detection rate is 0.01 on the restricted overall test (shaded area in Figure \ref{fig-over-18-det-alpha-4discs}), i.e.\ Recall = 0.99.
Precision (Pr) measures if detected minors are truly minors.
Recall (Re) measures whether all minors have been correctly detected.
F1 and F2 scores lie between the two measures. For each measure (column), the three experiments with the best values are highlighted.}\label{tab_quantitative}
\end{table}

In Table \ref{tab_quantitative}, we present quantitative measures at the discriminator threshold defined above for the restricted overall test (same as in Figure \ref{fig-over-18-det-alpha-4discs}).
The columns referring to the ASORES-39k restricted test reflect the ranking of the curves in the gray region in Figure \ref{fig-over-18-det-alpha-4discs}, and the best accuracy is achieved in the second-last row \emph{\ModelD}. It achieves an F2 score of 0.875 on the Restricted Overall Test and 0.813 on the ASWIFT-20k wild test, both at a fixed recall of 0.99. This represents an absolute improvement of +0.074  and +0.071 in F2, respectively, over the baseline age estimator \emph{model 0}.
The columns for the ASWIFT-20k test in the same table prove that the reasonable accuracy of this model carries nicely over to unconstrained conditions. In fact, \emph{\ModelD{}} is the only model that can maintain the Recall value of 0.99 in the Wild test.

\subsection{Performance across age thresholds and effects on age estimation}
\label{sec_experiments_12_1518}

We now evaluate whether adding multi-task heads benefits detection at the other relevant age thresholds (12, 15, and 21). In addition, we examine how these multi-task heads influence age regression performance.

\subsubsection{Impact of the training set, training scheme, and heads}

In Table \ref{tab_quantitative_all_levels}, the quantitative analysis is extended to the underage detection tasks, with age levels 12, 15, 18, and 21 years, quantified by the F2 score.
\begin{table}[htb]
\resizebox{\linewidth}{!}{\begin{tabular}{cclrrrrrrrrrrc}
\toprule
 &  &  & \multicolumn{5}{c}{ASORES-39k (restricted test)} & \multicolumn{5}{c}{ASWIFT-20k (unconstrained)} &   \\
 &  &   & F2 & F2  & F2 & F2  & MAE & F2  & F2  & F2 & F2 & MAE & \\
Heads & CACD & res & $T\!=\!12$ & $T\!=\!15$ & $T\!=\!18$ & $T\!=\!21$ & (years) & $T\!=\!12$ & $T\!=\!15$ & $T\!=\!18$ & $T\!=\!21$ & (years)  & \\
\midrule
\multirow[c]{4}{*}{\labelA} & \multirow[c]{2}{*}{with} & no & 0.748 & 0.752 & 0.801 & 0.846 & 4.175 & 0.666 & 0.689 & 0.742 & 0.830 & 4.818 & (0) \\
 &  & yes & 0.818 & 0.775 & 0.812 & 0.854 & 4.145 & 0.751 & 0.718 & 0.757 & 0.831 & 4.836 &  \\
 & \multirow[c]{2}{*}{without} & no & 0.905 & 0.831 & 0.833 & 0.853 & 4.034 & 0.905 & 0.757 & 0.764 & 0.838 & \bfseries 4.441 &  \\
 &  & yes & 0.843 & 0.815 & 0.833 & 0.854 & 4.174 & 0.781 & 0.748 & 0.770 & 0.836 & 4.688 &  \\
\midrule
\multirow[c]{4}{*}{\labelB} & \multirow[c]{2}{*}{with} & no & 0.846 & 0.800 & \bfseries 0.869 & 0.861 & \bfseries 4.022 & 0.778 & 0.728 & \bfseries 0.817 & 0.842 & 4.603 & (A) \\
 &  & yes & 0.846 & 0.783 & 0.868 & 0.857 & 4.178 & 0.769 & 0.712 & 0.803 & 0.837 & 4.838 & (B) \\
 & \multirow[c]{2}{*}{without} & no & 0.847 & 0.771 & 0.848 & 0.841 & 4.052 & 0.786 & 0.695 & 0.785 & 0.832 & \bfseries 4.464 &  \\
 &  & yes & 0.926 & 0.847 & \bfseries 0.870 & 0.858 & 4.098 & 0.910 & 0.784 & 0.807 & 0.838 & 4.618 & (C) \\
\midrule
\multirow[c]{4}{*}{\labelC} & \multirow[c]{2}{*}{with} & no & \bfseries 0.960 & \bfseries 0.929 & 0.865 & \bfseries 0.885 & \bfseries 3.919 & \bfseries 0.940 & \bfseries 0.897 & \bfseries 0.811 & \bfseries 0.857 & 4.501 &  \\
 &  & yes & 0.939 & 0.826 & 0.863 & \bfseries 0.875 & 4.173 & 0.846 & 0.776 & 0.787 & \bfseries 0.847 & 4.807 &  \\
 & \multirow[c]{2}{*}{without} & no & \bfseries 0.961 & \bfseries 0.947 & 0.864 & 0.869 & \bfseries 3.977 & \bfseries 0.952 & \bfseries 0.918 & 0.796 & 0.843 & \bfseries 4.313 &  \\
 &  & yes & \bfseries 0.959 & \bfseries 0.938 & \bfseries 0.875 & \bfseries 0.875 & 4.062 & \bfseries 0.933 & \bfseries 0.909 & \bfseries 0.813 & \bfseries 0.851 & 4.468 & (D) \\
\bottomrule
\end{tabular}
}
\caption{Quantitative measurement of the underage detection (F2 score) at the other age levels $T = 12, 15, 18, 21$ and the age regression error measure MAE for the Restricted Overall Test and the Wild Test.
The variation points are the inclusion of heads for underage discrimination, whether to include CACD in the (overall) training set, and whether to use the age-equalized resampling.
Experiments use the weight-sharing MLP (denoted WS in Table \ref{tab_quantitative_all_levels_2} below).
For each measure (column), the top three experiments are highlighted.}\label{tab_quantitative_all_levels}
\end{table}
Several general observations can be made.
Higher F2 scores can be achieved for lower age thresholds (under-12-year F2 scores up to 0.96) than for higher age thresholds (under-21-year F2 scores below 0.89).
This is a property of the data and reflects the fact that facial features change more rapidly with age at young ages than at older ages; It is easier to distinguish under-12-year-olds from the rest of the population than, e.g., under-21-year-olds from the rest.
Further, as expected, the ASWIFT-20k test is harder than the ASORES-39k test, regarding both underage discrimination and age estimation.

The age-regression error improves relative to the baseline (model 0) in the first line as soon as one or several underage detectors are added.
This is related to three effects: (a) a synergistic effect between the related tasks, (b) a regularizing effect of an additional task avoiding overfitting, and (c) a dedicated underage detector is an advantage over using the age estimate for the underage detection tasks. The underage detectors are beneficial for all tasks and in both the restricted and the wild test cases.

Among the settings for the heads, the trend is that the combination of removing CACD and age-equalized sampling leads to superior accuracy.
In summary, to this point, we have developed two techniques to improve on age estimation and underage detection with respect to the baseline (mark (0) in Table \ref{tab_quantitative_all_levels}): \textbf{(1)} Adding underage discriminators has a direct benefit for underage detection, but is also beneficial for age estimation (synergy of related tasks, regularizing effect); \textbf{(2)} Curate the training data by removing noisy age labels and age-equalization.

These results confirm that multi-threshold detection heads - \emph{\ModelD{}} - improve classification performance across age boundaries and enhance age regression accuracy.

\subsubsection{Effect of architecture and loss tuning on detection and estimation}\label{sec_mlp}
Above, we have shown that the best results for age estimation and underage detection in the wild are achieved with a FaRL + MLP model with four underage detectors, trained on the cleaned overall age dataset (without CACD) in an age-equalized manner.
However, there are several further variation points to be investigated, namely the MLP architecture and three loss-function parameters: the class balancing parameter $\alpha$, the focusing parameter $\gamma$, and the age gap $G$.
Table \ref{tab_quantitative_all_levels_2} collects the performance measures of a collection of models that vary these factors.

\begin{table}[htb]
    \resizebox{\linewidth}{!}{\begin{tabular}{ccccrrrrrrrrrrc}
 \toprule
 &  & & & \multicolumn{5}{c}{ASORES-39k (restricted test)} & \multicolumn{5}{c}{ASWIFT-20k (unconstrained)} &   \\
 &  &  &   & F2 & F2  & F2 & F2  & MAE & F2  & F2  & F2 & F2 & MAE & \\
MLP & $G$ & $\gamma$ & $\alpha$ & $T\!=\!12$ & $T\!=\!15$ & $T\!=\!18$ & $T\!=\!21$ & (years) & $T\!=\!12$ & $T\!=\!15$ & $T\!=\!18$ & $T\!=\!21$ & (years) & \\
\midrule
\multirow[c]{7}{*}{WS} & 0 & 0 &  & 0.959 & 0.938 & \bfseries 0.875 & 0.875 & 4.062 & 0.933 & 0.909 & 0.813 & 0.851 & \bfseries 4.468 & (D) \\
 & 0 & 0 & yes & 0.961 & 0.915 & 0.869 & \bfseries 0.878 & 4.094 & 0.935 & 0.879 & 0.804 & 0.850 & 4.631 &  \\
 & 0 & 2 &  & 0.954 & 0.922 & 0.867 & 0.872 & 4.083 & 0.905 & 0.876 & 0.805 & 0.848 & 4.498 &  \\
 & 1 & 0 &  & 0.957 & \bfseries 0.954 & \bfseries 0.879 & 0.877 & \bfseries 4.044 & 0.931 & 0.911 & 0.810 & 0.852 & \bfseries 4.473 &  \\
 & 2 & 0 &  & 0.958 & 0.947 & 0.873 & 0.873 & 4.060 & 0.944 & 0.903 & 0.778 & \bfseries 0.854 & 4.491 &  \\
 & 3 & 0 &  & 0.949 & 0.934 & 0.854 & 0.874 & 4.063 & 0.917 & 0.884 & 0.819 & 0.847 & 4.488 &  \\
 & rel & 0 &  & 0.960 & 0.946 & 0.856 & 0.873 & \bfseries 4.034 & \bfseries 0.948 & 0.911 & \bfseries 0.822 & 0.846 & \bfseries 4.450 & (E) \\
\midrule
I-512 & 0 & 0 &  & \bfseries 0.962 & 0.941 & \bfseries 0.876 & \bfseries 0.883 & \bfseries 4.041 & 0.946 & \bfseries 0.913 & 0.801 & 0.852 & 4.572 &  \\
\midrule
\multirow[c]{4}{*}{I-256} & 0 & 0 &  & \bfseries 0.965 & \bfseries 0.948 & 0.869 & \bfseries 0.878 & 4.065 & 0.946 & \bfseries 0.918 & 0.809 & \bfseries 0.853 & 4.520 &  \\
 & 3 & 0 &  & 0.961 & 0.931 & 0.864 & 0.872 & 4.081 & \bfseries 0.947 & 0.881 & \bfseries 0.840 & 0.851 & 4.496 &  \\
 & 3 & 2 &  & 0.931 & 0.921 & 0.827 & 0.876 & 4.075 & 0.904 & 0.884 & 0.810 & 0.848 & 4.556 &  \\
 & rel & 0 &  & \bfseries 0.965 & \bfseries 0.951 & 0.857 & 0.875 & 4.068 & \bfseries 0.955 & \bfseries 0.916 & \bfseries 0.833 & \bfseries 0.854 & 4.475 & (F) \\
\midrule
I-128 & 0 & 0 & & 0.958 & 0.937 & 0.868 & 0.866 & 4.075 & 0.920 & 0.904 & 0.800 & 0.845 & 4.504 &  \\
\bottomrule
\end{tabular}
}
    \caption{Performance measures of underage detection at age thresholds $12, 15, 18, 21$, and age regression.
    Common parameters: heads = \emph{age + 12,15,18,21}, training without CACD and with resampling (like model D in Table \ref{tab_quantitative_all_levels}).
    Variation points are the architecture of the MLP [weight-sharing MLP (WS) vs.\ two-layer independent MLP with intermediate width $m$ (I-$m$), see Sec.\ \ref{sec_MultiAge}], the age gap $G$ (``rel'' means age gap according to Eq.\ \ref{eq_rel_age_gap}), the focusing parameter $\gamma$, and the class reweighting $\alpha$.
    Top three values in bold face.}\label{tab_quantitative_all_levels_2}
\end{table}

As seen in Figure \ref{fig-over-18-training-set-det} and Table \ref{tab_quantitative} for under-18 discrimination, the $\alpha$ parameter does not improve accuracy for the other age thresholds either, if resampling is in place.
Two experiments with the gamma parameter increased to $\gamma=2$ are shown (3rd vs.\ 2nd line in the WS-block; 3rd line vs.\ 2nd line in the I-256 block in Table \ref{tab_quantitative_all_levels_2}), where the focal loss (Eq.\ \ref{eq:alpha_FocalLoss}) focuses on the most ``difficult'' patterns. This leads to no clear improvement either.
Our hypothesis for this failure is that focusing on the difficult examples may lead to overfitting to edge cases and label noise.

The next variation point, introducing an age gap, is motivated by the opposite approach, namely by \emph{removing} edge cases from the loss function (Section \ref{sec_age_gap}). This adds the configuration of each discriminator to the configuration space. We have tried fixed age gaps of ($G=1,2,3$, i.e.\ with age gap $[-G, G]$ with respect to the age threshold) for all four discriminators.
The data for MLP = WS and $G=0,1,2,3$ in Table \ref{tab_quantitative_all_levels_2} does not show clear trends on ASORES-39k, but it does on ASWIFT-20k:
Underage detection benefits from the age gap, but not at the same gap parameter: the F2 of $T=12$ increases from 0.933 to 0.944 at $G=2$; the F2 of $T=18$ increases from 0.813 to 0.819 at $G=3$.
Thus, the optimal value of $G$ seems to scale with the age threshold $T$, as already motivated in Section \ref{sec_age_gap}, Eq. \ref{eq_rel_age_gap}.
This configuration leads to the model labeled as (E) in Table \ref{tab_quantitative_all_levels_2}.
It shows improvements over model D for under-12, under-15 detection and age regression, but not in under-18 and under-21 detection.

\paragraph{MLP architecture}
In the lower part of Table \ref{tab_quantitative_all_levels_2}, we show the results of models where this weight-sharing two-layer MLP is replaced with a regular MLP of two (independent) linear layers, each followed by a ReLU activation function.
Input and output dimensions are kept at 512, but the intermediate width is varied to adjust the capacity of the MLP.
This ``bottleneck'' is to make up for the regularizing effect of the shared weights in the original SW implementation.
The results show that the MLP with an intermediate bottleneck (I-256) achieves slightly better underage detection F2 scores, but is worse in age estimation. If combined with the gap of Eq.\ \ref{eq_rel_age_gap} ($G$ = rel), however, the resulting model F improves consistently over model D on the ASWIFT-20k test.

In this Subsection \ref{sec_mlp}, we have shown that the focal-loss parameters $\alpha$ and $\gamma$ have only minor effects and cannot improve the performance of model D.
The gap parameter, in turn, has the potential to improve it, if configured correctly.
Finally, we have shown that the MLP of independent linear layers and a bottleneck of width 256 can replace the weight-sharing MLP, reaching comparable accuracy.
Combined with the age gap (model F), it consistently outperforms model D in the ASWIFT-20k wild test, albeit with slight degradation in the ASORES-39k test for under-18 detection and age regression.
Still, a model with more even accuracy across domains is desirable.

\subsection{Ablation study and comparison with baselines}
After designing age estimation and minor detection models based on the frozen FaRL backbone, in particular the model F listed in Table~\ref{tab_quantitative_all_levels_2}, three questions remain to be answered. (a) Is the frozen backbone the best choice, or can further accuracy be gained by fine-tuning? (b) Given the composed training set and the resampling scheme, how important is the architecture of the age estimator and the minor-detection heads? (c) How does our model perform compared to state-of-the-art methods?

\subsubsection{Age estimation}

Table \ref{tab_age_fine} shows the mean absolute regression error of several experiments addressing these questions. The first line, \emph{Age + 4 $\times$ Minor, no fine-tuning}, is our Model F from above.
\begin{table}[bt]
\small\centering
\newcommand{\finetune}[1]{
\ifthenelse{\equal{#1}{0}}{no fine-tuning}{}%
\ifthenelse{\equal{#1}{1}}{from 10$^\mathrm{th}$ block}{}%
\ifthenelse{\equal{#1}{2}}{from 5$^\mathrm{th}$ block}{}%
\ifthenelse{\equal{#1}{3}}{full backbone}{}%
\ifthenelse{\equal{#1}{4}}{from start}{}%
}
\newcommand{\baselineDEX}{DEX: VGG-16, IMDB-WIKI. \cite{Rothe_2015_DEX}}
\newcommand{\baselineMiVOLO}{MiVOLO, vision transformer, \cite{Kuprashevich_2024_MiVOLO}}
\newcommand{\agefourm}{$\mathrm{Age} + 4 \times \mathrm{minor}$}
\begin{tabular}{llccccc}
\toprule
             &             & ASORES & ASORES & ASWIFT & ASWIFT & \\
Prediction heads & Fine-tuning &   39k  &   OOD\,18k  &  20k   &  OOD\,10k& \\
\midrule
\agefourm & \finetune{0} & 4.068 & \bfseries 4.212 & \bfseries 4.475 & \bfseries 4.807 & (F) \\
  & \finetune{1} & \bfseries 3.925 & \bfseries 4.146 & 4.527 & 5.038 & \\
  & \finetune{2} & \bfseries 3.946 & 4.220 & 4.586 & 5.204 & \\
  & \finetune{3} & \bfseries 3.990 & 4.220 & 4.628 & 5.255 & \\
\midrule
Age only & \finetune{0} & 4.054 & 4.228 & \bfseries 4.460 & \bfseries 4.805 & \\
  & \finetune{1} & 4.011 & 4.266 & 4.642 & 5.292 & \\
  & \finetune{2} & 4.069 & 4.562 & 4.638 & 5.440 & \\
\midrule
Direct regression & \finetune{0} & 4.049 & \bfseries 4.082 & 4.483 & \bfseries 4.703 & \\
  & \finetune{1} & 4.005 & 4.254 & \bfseries 4.461 & 4.886 & \\
  & \finetune{2} & 4.169 & 4.845 & 4.733 & 5.730 & \\
\midrule
\multicolumn{2}{l}{\baselineDEX} & 6.577 & 7.844 & 7.632 & 9.323 & \\
\multicolumn{2}{l}{\baselineMiVOLO} & 4.560 & 4.355 & 5.653 & 5.738 & \\
\bottomrule
\end{tabular}

\caption{Fine-tuning and comparison to two baseline methods (MAE in years).
The out-of-distribution (OOD) variants of the tests are restricted to the datasets CACD2000, FG-NET, CASIA-Face-Africa, and Dartmouth, i.e., those not used in the overall training set, and images where the MiVOLO face detector failed were excluded, allowing for a fair comparison with the external baseline models DEX and MiVOLO.}\label{tab_age_fine}
\end{table}%
We compare the FaRL backbone with three different setups for the prediction heads (age + 4 $\times$ minor detection; age only; direct age regression) and further vary the degree of fine-tuning of the backbone (no fine-tuning; fine-tuning from the 10$^\mathrm{th}$ resblock on; fine-tuning from the 5$^\mathrm{th}$ resblock on; fine-tuning of the entire vision transformer).
The impact of these changes on the prediction accuracy depends on the test case: Sometimes, a bit of fine-tuning helps, but not always:
For the ASORES test, moderate fine-tuning (from the 10$^\mathrm{th}$ resblock onward) further reduces the estimation error in all three setups. 
However, ASWIFT and ASWIFT-OOD tests are best without fine-tuning. The differences between the losses (age + 4 $\times$ minor detection, age only, and direct regression) are not significant. 
Since the main motivation of our paper is robust performance under adverse conditions, we prioritize the ASWIFT test. The conclusion is that neither fine-tuning nor changes to the loss function can significantly improve age-regression accuracy, at least with the current training data.

The last block of the table shows the mean absolute errors achieved by the classic DEX model \cite{Rothe_2015_DEX} and the state-of-the-art MiVOLO model \cite{Kuprashevich_2024_MiVOLO}. For a fair comparison, consider the columns ASORES-OOD and ASWIFT-OOD.
Our model “F” outperforms both baseline models.
While the MiVOLO model is close to ours on the ASORES-OOD test, our model has a significant advantage on the ASWIFT-OOD test.

\paragraph{Multi-task vs.\ single-task}
To evaluate the synergy of the tasks of age estimation and underage detection, we compare the model (F) to a single-task underage detector with the same parameters. If the age gap is preserved, the single-task model fails: the separation of adults and minors is so bad that no recall better than 0.92 can be reached.
When the age gap is removed, this failure is healed, and the model reaches reasonable accuracy again, with a performance similar to Model F on ASORES-39k. On the ASWIFT-20k test, however, we measure
Pr = 0.439, F2 = 0.791, 
which is worse than the multitask model F (Pr = 0.512, F2 = 0.833, see also Table \ref{tab_quantitative_all_levels_2}).
Thus, the age gap works well only in the multitask setup, last hidden layers are shared with the age regressor, 
but then, this setup leads to enhanced generalization.

\subsubsection{Underage detection and age classification with partially occluded faces}
\begin{table}[bt]
\centering\small
\begin{tabular}{lcc}
  \toprule
  Model & MAE & Classification \\
  \midrule
  Chaves et al.\ \cite{Chavez_VISAPP2020}, (IMDB, 4000 img/age, only original) & \unit[5.02]{y} & Acc = 33.9\% \\
  Chaves et al.\ \cite{Chavez_VISAPP2020}, (IMDB, 4000 img/age, original + occluded) & \unit[4.07]{y} & Acc = 39.2\% \\
  Model F \textbf{(ours)} & \unit[4.11]{y} & Acc = 47.6\% \\
  \bottomrule
\end{tabular}
\caption{Performance comparison with Cháves et al.\ on the occluded + original test of Ref.\ \cite{Chavez_VISAPP2020}.
}\label{tab_comp_visapp}
\end{table}%
We have evaluated our model F on the test set of \cite{Chavez_VISAPP2020}, which is an age-balanced dataset ranging from 0 to 25 years with 1000 images per year and composed of images from
the \emph{Diversity in Faces} dataset \cite{Merler_2019_DIF} (86\%),
IMDB-WIKI \cite{Rothe_2018_IMDBWIKI} (5.2\%),
FG-NET \cite{Lanitis_2002_FGNET} (0.01\%),
UTKFace \cite{Zhang_2017_UTKFace} (5.8\%),
APPA-REAL  \cite{Agustsson_2017_AppaReal} (2\%), and
AgeDB \cite{Moschoglou_2017_AgeDB} (1.3\%).
Only the last three datasets (9.1\%) have a potential overlap with our training set since the exact train-test splits are not reproducible anymore.
Each image is tested twice, with and without a black bar drawn as an artificial occlusion over the eye region.
The figures of merit are the MAE measuring the regression error and the mean accuracy for classifying faces into five age groups with boundaries 0, 6, 11, 16, 18, and 26 years.
In Table \ref{tab_comp_visapp}, we compare our model F to the values reported in Ref. \cite{Chavez_VISAPP2020}.
Even though our model is not specifically trained for occluded faces, its age estimation accuracy ($\mathrm{MAE} = \unit[4.11]{y}$) comes very close to that of the strongest model of \cite{Chavez_VISAPP2020}, trained with original and occluded images ($\mathrm{MAE} = \unit[4.07]{y}$).
In terms of classification accuracy, model F clearly outperforms all models of Chaves et al.\ \cite{Chavez_VISAPP2020}.

\subsection{Analysis of demographic effects and failure modes}
In this section, we analyze in more detail how our model F responds to shifts in the data distribution and under what conditions it is likely to fail.
\begin{figure}[b]
\centering
\includegraphics[width=0.95\linewidth,trim=0 20 0 20]{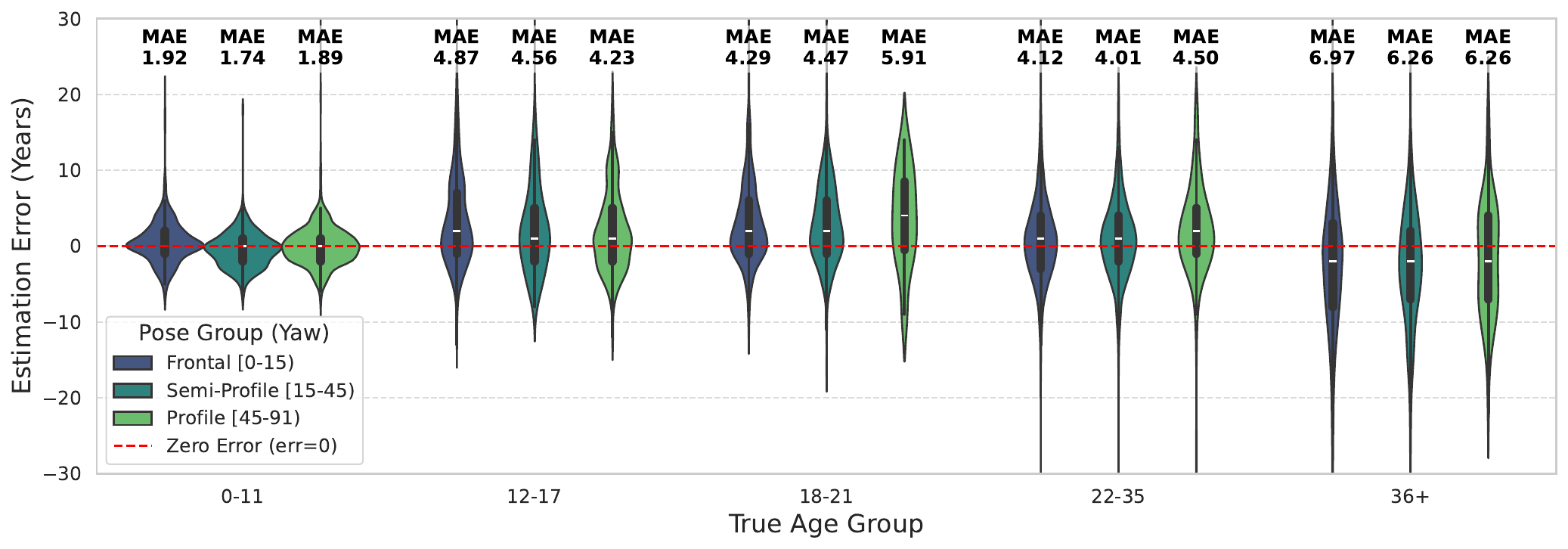}
\caption{Error distribution by age and yaw angle on the ASWIFT-20k test.}\label{fig_err_age_yaw}
\end{figure}%
In Figure \ref{fig_err_age_yaw} we show the distribution of the age estimation error, divided into groups of age and yaw angle. Within each age group, the MAE and the distribution of the error vary only very little with the yaw angle, highlighting the robustness of the underlying FaRL backbone. We do observe, however, the well-known pattern that age estimation in terms of absolute regression errors is easier at young ages than at older ages.

\begin{table}[h!]
    \resizebox{\linewidth}{!}{\begin{tabular}{lccccccccccccr}
\toprule
 & \multicolumn{3}{c}{Under-12 Detection} & \multicolumn{3}{c}{Under-15 Detection} & \multicolumn{3}{c}{Under-18Detection} & \multicolumn{3}{c}{Under-21 Detection} &  \\
Test & Pr & F1 & Re & Pr & F1 & Re & Pr & F1 & Re & Pr & F1 & Re & MAE \\
\midrule
Source: ASORES-39k & 0.88 & 0.93 & 0.99 & 0.82 & 0.90 & 0.99 & 0.56 & 0.71 & 0.99 & 0.60 & 0.75 & 0.99 & 4.07 \\
Source: ASWIFT-20k & 0.83 & 0.90 & 0.99 & 0.73 & 0.83 & 0.98 & 0.51 & 0.67 & 0.99 & 0.56 & 0.71 & 0.98 & 4.48 \\
\midrule
Gender: Male & 0.90 & 0.94 & 0.99 & 0.81 & 0.89 & 0.99 & 0.55 & 0.70 & 0.98 & 0.56 & 0.71 & 0.98 & 3.94 \\
Gender: Female & 0.80 & 0.89 & 0.99 & 0.70 & 0.81 & 0.98 & 0.45 & 0.62 & 0.99 & 0.53 & 0.69 & 0.99 & \itshape 4.71 \\
Race: Asian & 0.89 & 0.94 & 1.00 & 0.63 & 0.77 & 0.99 & \itshape 0.26 & \itshape 0.40 & \itshape 0.94 & \itshape 0.36 & \itshape 0.53 & 0.99 & 3.88 \\
Race: White & 0.87 & 0.93 & 0.99 & 0.85 & 0.91 & 0.99 & 0.61 & 0.76 & 1.00 & 0.68 & 0.80 & 0.99 & 4.17 \\
Race: Black & \itshape 0.29 & \itshape 0.44 & \itshape 0.95 & \itshape 0.14 & \itshape 0.24 & \itshape 0.90 & \itshape 0.24 & \itshape 0.38 & \itshape 0.94 & \itshape 0.42 & \itshape 0.59 & \itshape 0.97 & 4.50 \\
Pose: Extreme Yaw & 0.86 & 0.93 & 1.00 & 0.93 & 0.96 & 0.99 & 0.92 & 0.96 & 1.00 & 0.92 & 0.96 & 1.00 & 2.92 \\
Pose: Extreme Pitch & 0.88 & 0.93 & 1.00 & 0.74 & 0.85 & 0.99 & 0.47 & 0.64 & 0.99 & 0.56 & 0.71 & 0.99 & 4.47 \\
\midrule
Image: Dark & 0.77 & 0.86 & 0.98 & 0.67 & 0.80 & 0.97 & 0.49 & 0.65 & 0.98 & 0.52 & 0.68 & 0.98 & 4.36 \\
Image: Blurry & 0.78 & 0.87 & 0.98 & 0.52 & 0.68 & 0.98 & 0.31 & 0.47 & \itshape 0.97 & \itshape 0.39 & \itshape 0.55 & \itshape 0.97 & \itshape 4.66 \\
Image: Low Contrast & 0.74 & 0.85 & 0.98 & \itshape 0.43 & \itshape 0.59 & 0.96 & 0.28 & 0.44 & 0.97 & 0.43 & 0.59 & \itshape 0.96 & \itshape 4.70 \\
Image: Low Saturation & \itshape 0.68 & \itshape 0.80 & \itshape 0.96 & \itshape 0.32 & \itshape 0.47 & \itshape 0.96 & \itshape 0.21 & \itshape 0.35 & \itshape 0.93 & \itshape 0.40 & \itshape 0.57 & 0.98 & 4.41 \\
Image: High Saturation & \itshape 0.72 & \itshape 0.83 & 0.99 & 0.64 & 0.77 & \itshape 0.96 & 0.49 & 0.65 & 0.99 & 0.54 & 0.70 & 0.99 & 4.62 \\
\midrule
Expr: High Arousal & 0.89 & 0.94 & 0.99 & 0.85 & 0.92 & 0.99 & 0.65 & 0.78 & 0.99 & 0.67 & 0.80 & 0.99 & 4.11 \\
Expr: Low Valence & 0.83 & 0.90 & 0.99 & 0.82 & 0.90 & 0.99 & 0.60 & 0.75 & 0.99 & 0.63 & 0.77 & 0.99 & 4.21 \\
Expr: High Valence & 0.89 & 0.94 & 0.99 & 0.79 & 0.88 & 0.99 & 0.49 & 0.65 & 0.99 & 0.57 & 0.72 & 0.98 & 4.50 \\
\midrule
DB: CASIA-Face-Africa & \itshape 0.11 & \itshape 0.19 & \itshape 0.89 & \itshape 0.10 & \itshape 0.18 & \itshape 0.81 & \itshape 0.21 & \itshape 0.34 & 0.98 & 0.50 & 0.66 & \itshape 0.94 & \itshape 7.22 \\
DB: FG-NET & 0.86 & 0.91 & \itshape 0.96 & 0.79 & 0.88 & 0.99 & 0.78 & 0.88 & 0.99 & 0.88 & 0.93 & 0.98 & 3.63 \\
\bottomrule
\end{tabular}
}
    \caption{Performance measures of Model F across demographic and image-condition subsets. The top rows show the ASORES-39k and ASWIFT-20k tests (see also the rows marked with (F) in Tables \ref{tab_quantitative_all_levels_2} and \ref{tab_age_fine}). Further rows show subsets of the union of ASORES-39k and ASWIFT-20k. Gender and Race: estimates of the FairFace model \cite{Karkkainen_2021_FairFace}; Pose: highest 10\% of the InsightFace angles; Image properties: highest/lowest 20\% of the image properties; Expression: lowest/highest 10\% of the arousal and valence parameter of \cite{Vonikakis_2021_expression} (details Sec.\ \ref{sec_selection}).
    The last rows show the \emph{CASIA-Face-Africa} subset (3375 images) of ASWIFT-20k  and the \emph{FG-NET} dataset (1k images).
    In each column, the four worst-performing cases are highlighted in italics.}\label{tab-performance-breakdown}
\end{table}

A more extensive analysis is given in Table \ref{tab-performance-breakdown}, which lists performance measures, both for underage detection and for age regression, for different subsets of the union of ASORES-39k and ASWIFT-20k.
In most cases, the recall (Re) remains high and the precision (Pr) does not drop too much. In particular, there are no problems with large pose angles and strong expressions.
However, we do observe relevant deterioration of performance for
the Female group (age regression), Asian and Black groups (problems with false adult detections, but recall stays high), and the Blurry, Low Contrast, and Low Saturation groups (problems with both underage detection and age regression).

These observations should be interpreted cautiously, as the test data is unevenly distributed. Females are strongly represented in CACD2000, while males are overrepresented in MORPH.
Asians mostly come from the AFAD dataset, and Blacks from MORPH and CASIA-Face-Africa, both of lower image quality.


The last two rows show two cases of cross-dataset validation, i.e., data from datasets that were not used in the training at all, not just hold-out sets.
The row CASIA-Face-Africa shows results of the 3375 lowest quality webcam images from the CASIA-Face-Africa dataset selected for the ASWIFT-20k test (see Table \ref{wild-test-composition}). In this doubly difficult case, the model reaches its limits; the precision breaks down, but the recall remains high.
The results measured with the FG-NET dataset show that the MAE of $3.63$ years is better than the values in Tables \ref{tab_age_benchmark} and \ref{tab_age_overall} and values reported elsewhere (Paplhám \& Franc \cite{Paplham_CVPR2024_benchmark} report $4.41$ years for a pre-trained regression model trained on UTKFace).

\section{Conclusions} \label{sec_conclusions}

In this work, we propose a multitask architecture for facial age estimation and underage detection that is robust in unconstrained imagery. The model performs age regression and binary underage classification at four critical legal thresholds (12, 15, 18, and 21 years) simultaneously.
%
The model leverages the face representation of the FaRL backbone and is trained on a curated multi-source dataset, excluding the noisy CACD2000 dataset, and by removing label noise. To address the imbalance in age distribution, we propose a resampling strategy in which underrepresented age groups (minors) are sampled more frequently in the training set than the others. This approach is superior to adding weight factors ($\alpha$) in the loss function.
In conjunction with the age-estimation task in the multi-task architecture, removing edge cases from the binary loss functions by introducing an ``age gap'' improves the generalization of the trained model.
Thus, the multi-task setup is able to exploit synergies between the related tasks, age-regression and underage detection.

For evaluation, we designed a challenging benchmark, the ASWIFT-20k wild test, which deliberately includes images with extreme head poses, facial expressions, and illumination conditions.
We have analyzed our model's performance on subsets of the test data, defined by head pose, facial expression, demographic groups, and imaging conditions.
On the plus side, our model is robust to pose and expression variation and shows lower estimation errors for young subjects than for older ones.
There are weaknesses, however, for Black subjects and under difficult imaging conditions.

More investigation is needed to identify and remedy such biases.
Expanding the training set with additional age-labeled images for underrepresented combinations of age, ethnicity, gender, and image conditions could be essential.
Generative data augmentation techniques (e.g., facial expression transfer) may address demographic gaps while maintaining diversity and realism.
Low-quality images could be enhanced by applying super-resolution techniques that preserve the age-relevant features of the image.
Additionally, we intend to refine the ASWIFT-20k benchmark for a more balanced distribution, reducing correlations between parameters like \textit{pose}, \textit{race}, or \textit{age}, allowing a more rigorous and fair evaluation of model robustness.

Finally, we acknowledge the ethical considerations related to the use of facial analysis technology.
While our research is motivated by its application to minor protection, specifically to CSEM detection, we recognize that age estimation tools could be misused.
Scenarios such as pervasive surveillance and automated discrimination based on age are significant concerns. Therefore, we stress the responsible deployment of these systems. Their use should be limited to specific, ethically evaluated, and legally sanctioned cases.
Moreover, given the identified bias and performance gaps, we emphasize the importance of human supervision in any critical deployment to grant fairness, correct errors and prevent harm.

The code of this work is released at \url{https://gitlab.com/gvis-lab/MultiAge}.

\section*{Acknowledgements}
This work was funded by the Recovery, Transformation, and Resilience Plan, of the European Union (Next Generation)
through the LUCIA project (Fight against Cybercrime by applying Artificial Intelligence) granted by INCIBE to University of León.

\renewcommand{\url}[1]{\href{#1}{URL}}
\providecommand{\URLprefix}{}

\bibliographystyle{elsarticle-num-names}
\bibliography{references}

\end{document}